\documentclass[]{fairmeta}
\usepackage{amssymb}
\usepackage{multirow}
\usepackage{bigdelim}
\usepackage{todonotes}
\usepackage{longtable}
\usepackage{tabularray}
\usepackage{wrapfig}
\usepackage[most]{tcolorbox} 
\usepackage{xcolor}
\usepackage{url}

\usepackage[utf8]{inputenc}
\usepackage{tikz}
\usepackage{amsmath}
\usepackage{caption}
\usepackage{subcaption}
\usepackage[table]{xcolor}

\usepackage{pifont}
\usepackage{makecell}
\usepackage{array} 
\usepackage{makecell}
\usepackage{ulem}

\usepackage{siunitx}
\makeatletter
\newcommand{\onedot}{\@ifnextchar.{}{.}}
\makeatother

\def\eg{\emph{e.g}\onedot}

\usetikzlibrary{shapes.geometric, arrows.meta, positioning, decorations.pathmorphing}

\definecolor{predblue}{RGB}{66,153,225}
\definecolor{gtgreen}{RGB}{72,187,120}
\definecolor{matchgray}{RGB}{160,174,192}
\definecolor{textgray}{RGB}{74,85,104}
\definecolor{titlegray}{RGB}{45,55,72}

\tikzset{
    mydiamond/.style={
        diamond,
        fill=#1,
        draw=#1,
        minimum size=6pt,
        inner sep=0pt
    }
}

\newtcolorbox{promptbox}[1][]{
    breakable,                       
    colback=gray!5!white,            
    colframe=gray!75!black,          
    enhanced,                        
    arc=3pt,                         
    boxrule=0.5pt,                   
    left=5pt, right=5pt, top=5pt, bottom=5pt, 
    #1 
}

\usepackage{siunitx}
\definecolor{prompt}{HTML}{5f84e4}
\definecolor{img}{HTML}{820100}

\setcitestyle{numbers,square}

\title{OmniScript: Towards Audio-Visual Script Generation for Long-Form Cinematic Video}

\author[*]{Junfu Pu}
\author[*]{Yuxin Chen}
\author[*]{Teng Wang}
\author[]{Ying Shan}

\affiliation[]{ARC Lab, Tencent}

\contribution[*]{Equal Contribution}

\abstract{Current multimodal large language models (MLLMs) have demonstrated remarkable capabilities in short-form video understanding, yet translating long-form cinematic videos into detailed, temporally grounded scripts remains a significant challenge. This paper introduces the novel video-to-script (V2S) task, aiming to generate hierarchical, scene-by-scene scripts encompassing character actions, dialogues, expressions, and audio cues. To facilitate this, we construct a first-of-its-kind human-annotated benchmark and propose a temporally-aware hierarchical evaluation framework. Furthermore, we present \textit{OmniScript}, an 8B-parameter omni-modal (audio-visual) language model tailored for long-form narrative comprehension. OmniScript is trained via a progressive pipeline that leverages chain-of-thought supervised fine-tuning for plot and character reasoning, followed by reinforcement learning using temporally segmented rewards. Extensive experiments demonstrate that despite its parameter efficiency, OmniScript significantly outperforms larger open-source models and achieves performance comparable to state-of-the-art proprietary models, including Gemini 3-Pro, in both temporal localization and multi-field semantic accuracy.
}

\date{April 13, 2026}

\metadata[Project Page]{\url{http://arcomniscript.github.io}}
\metadata[Code]{\url{https://github.com/TencentARC/OmniScript}}

\begin{document}

\maketitle

\begin{figure}
    \centering
    \vspace{0mm}  
    \includegraphics[width=1.0\textwidth]{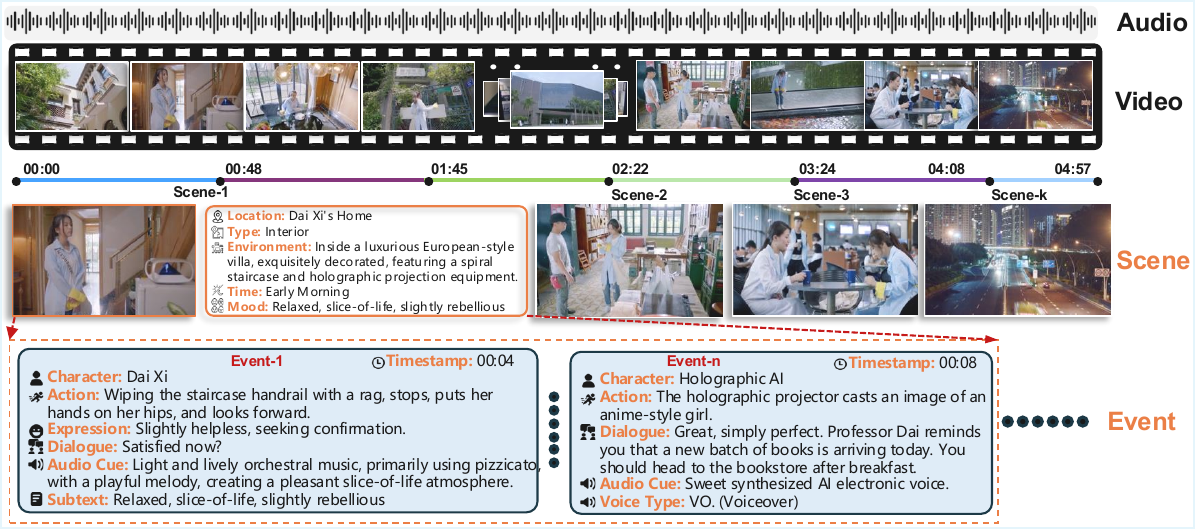}
    \caption{Overview of our Video-to-Script (V2S) framework. Given a long-form cinematic video, our pipeline performs temporally grounded scene-event parsing and generates a structured script with multimodal fields (dialogue, action, expression, and audio cues).}
  \label{fig:teaser} 
  \vspace{-6mm}
\end{figure}

\section{Introduction}
\label{sec:intro}

The analysis of cinematic content is a complex cognitive task, requiring the interpretation of a rich tapestry of visual and linguistic elements, including character relationships, narrative progression, and dialogue nuances. A deep understanding of these components is not only a cornerstone of computational media analysis but also holds significant practical value for the film and television industry, with potential applications in automated logging, content retrieval, and assisting human creativity. While recent advancements in multimodal large language models (MLLMs) \cite{liu2023visual, wang2025internvl3,zhu2025internvl3,deng2025emerging} have shown remarkable promise in video understanding, they predominantly focus on short-form video clips and tasks such as captioning or question-answering \cite{zhang2024llava,ge2025arc,shen2024longvu,yu2025minicpm}. The more ambitious and practical goal of translating a full-length movie or episode into a detailed, structured script remains a largely unexplored frontier.

In this paper, we introduce the first Video-to-Script (shown in~Fig.~\ref{fig:teaser}) multimodal large language model, a novel omni system designed to ingest long-form videos, ranging from several to tens of minutes, and generate a comprehensive, scene-by-scene script. 
This generated output includes not only the visual setting, such as the location, time of day, and atmosphere of each scene, but also a detailed breakdown of the in-scene narrative: character actions, dialogues (including voiceovers), and emotions. 
This task, which we term ``holistic script generation,'' pushes the boundaries of current video MLLMs by demanding fine-grained, long-range temporal understanding and coherent narrative synthesis.

However, the development of such a model is fraught with significant challenges. First, there is a critical lack of suitable training data. Annotating long-form videos with the necessary granularity, parsing complex multi-scene structures and intricate character interactions, is an exceptionally labor-intensive and time-consuming endeavor. This data scarcity poses a fundamental obstacle to training models capable of deep narrative comprehension. Second, evaluating the quality of generated scripts presents a unique metrological challenge. Unlike tasks with single ground-truth answers, the model's output is a long, open-ended, and richly structured narrative containing multiple elements with precise temporal stamps. Defining robust metrics that capture the accuracy, coherence, and completeness of such complex generations remains an open problem. Third, the autoregressive generation of such detailed descriptions is computationally prohibitive. Our preliminary analysis indicates that describing a mere two-minute clip requires approximately 4,000 tokens. As the video duration extends, the token count—and consequently the inference cost and time—can explode, creating a critical bottleneck for practical efficiency.

To overcome these hurdles, we propose a comprehensive and novel solution consisting of three key contributions. \textbf{First}, we introduce the first-of-its-kind movie and TV series understanding dataset. This dataset encompasses a diverse range of content, including both horizontal-format films and vertical-format short dramas, covering a wide array of themes and genres to ensure robust model training. \textbf{Second}, we establish a dedicated, human-annotated benchmark and a suite of evaluation metrics specifically designed for the video-to-script task. By rigorously assessing the performance of existing closed-source and open-source MLLMs, our benchmark not only quantifies the current state-of-the-art but also highlights the significant gap in research and capability that our work aims to fill. \textbf{Third}, we propose the first omni-modal (audio+visual) script generation model. 
We leverage extensive audio-visual pre-training, CoT-style SFT with plot and character relationship reasoning, and employ an open-ended reward for script quality. This reward is effectively utilized in a GRPO-based post-training phase to align the model with human preferences for coherent and accurate storytelling.

Extensive experiments demonstrate the superiority of our approach. Our model not only achieves performance comparable to 
 state-of-the-art closed-source models, including Gemini 3-Pro, on long-form video script generation tasks. This work represents a significant step toward automating the intricate process of video understanding and narrative generation, opening new avenues for research and application in computational media analysis.

\section{Related Work}
\label{sec:related_work}

\subsection{Cinematic Video Understanding and Narration}
Early cinematic understanding evolved from short-clip descriptions \cite{rohrbach2017movie, torabi2015using, rohrbach2015dataset,soldan2022mad} to plot-centric reasoning \cite{tapaswi2016movieqa, bain2020condensed, li2023ptvd} and structural annotations (e.g., AVA \cite{gu2018ava}, MovieGraphs \cite{vicol2018moviegraphs}, MovieNet \cite{huang2020movienet}). However, a persistent dichotomy remains: macro-plot analyses lack fine-grained temporal grounding, while structural annotations fail to yield readable, coherent narratives. Recent efforts like Movie101 \cite{yue2023movie101, yue2025movie101v2} generate role-aware narrations but still predominantly operate on isolated, dialog-free clips, failing to model full-length film intricacies. In contrast, our proposed Holistic Script Generation task requires transcribing full-length videos into temporally anchored, hierarchically structured scripts. Unlike existing datasets \cite{rohrbach2017movie, soldan2022mad, yue2023movie101} that entangle multimodal cues into coarse paragraph summaries, our annotation format strictly decouples fine-grained atomic elements, such as individual actions, expressions, dialogues, and audio cues, providing a much more rigorous benchmark for long-term narrative comprehension.

\subsection{Dense and Omni-Modal Video Captioning}
Traditional dense video captioning focuses on localizing salient events but typically yields sparse, concise summaries that neglect rich multimodal nuances \cite{krishna2017dense, geng2025longvale, pu2025arc}. Conversely, recent advancements in audio-visual captioning (e.g., AVoCaDO \cite{chen2025avocado}, video-SALMONN-2 \cite{tang2025video}, and DiaDem \cite{chen2026diadem}) achieve deep multimodal integration but generate holistic descriptions devoid of explicit temporal grounding. Empowered by advanced MLLMs, bridging this gap has emerged as a new frontier. Notably, a concurrent work, TimeChat-Captioner \cite{yao2026timechat}, introduces an "Omni Dense Captioning" task to generate timestamped, scene-level descriptions. While TimeChat-Captioner pioneers macro-scene structuring, its character actions, intents, and dialogues remain entangled within coarse summaries. Our OmniScript framework transcends these limitations by enforcing a strictly decoupled, hierarchical structure (\textit{Scene $\rightarrow$ Event $\rightarrow$ Field}) that explicitly isolates fine-grained atomic elements anchored to dense timestamps, effectively resolving the ambiguity present in prior captioning paradigms.

\vspace{-3mm}
\section{Video-to-Script Generation}
\label{sec:script_model}

In this section, we first define the cinematic video-to-script problem and output schema, then describe the memory-augmented automatic annotation pipeline used to construct high-quality training data.

\vspace{-2mm}
\subsection{Problem Definition}

Our task focuses on cinematic video understanding, where the goal is to transcribe a long-form movie/TV video into a temporally grounded, hierarchically structured script. Given an input video $V$ with duration $T$, the model predicts a structured output $\hat{Y}=\{\hat{M},\hat{S}\}$, where $\hat{M}$ denotes video-level metadata, and $\hat{S}$ denotes scene-event scripts.

\noindent\textbf{Input.}
The input is an untrimmed cinematic video $V$ containing complex narrative transitions, recurring characters, and multimodal cues (visual, speech, sound effects, and background music).

\noindent\textbf{Output Schema.}
The output follows a JSON-style schema with three levels:
\begingroup
\setlength{\leftmargini}{1.2em}
\begin{itemize}
    \item \textbf{Meta-level ($\hat{M}$):} global attributes, including title/description, total duration, and a character list.
    \item \textbf{Scene-level ($\hat{S}$):} a sequence of scenes $\{s_i\}_{i=1}^{N_s}$, where each scene contains a scene identifier, location, and coarse time attribute (e.g., day/night).
    \item \textbf{Event-level ($\hat{E}_i$):} each scene $s_i$ contains ordered events $\{e_{i,j}\}_{j=1}^{N_i}$ with timestamp and character identity, while each event includes at least one content field from \{dialogue, action, expression, audio cue (\eg sound event or BGM)\}.
\end{itemize}
\endgroup

Formally, for an event $e_{i,j}$ occurring at time $\tau_{i,j}$, we define
\begin{equation}
e_{i,j} = (\tau_{i,j}, c_{i,j}, d_{i,j}, a_{i,j}, x_{i,j}, u_{i,j}),
\end{equation}
where $c_{i,j}$ is character (or \texttt{Environment}), and $d/a/x/u$ denote dialogue/action/ expression/audio cue, respectively.

\noindent\textbf{Objective.}
The overall objective is to learn a mapping $f_\theta: V \rightarrow \hat{Y}$ that jointly optimizes: (1) temporal localization accuracy (when events happen), (2) character-consistent semantic parsing (who does/says what), and (3) multimodal narrative faithfulness (how the event is expressed through language, behavior, expression, and sound).

\begin{figure}[t]e
    \centering
    \includegraphics[width=\linewidth]{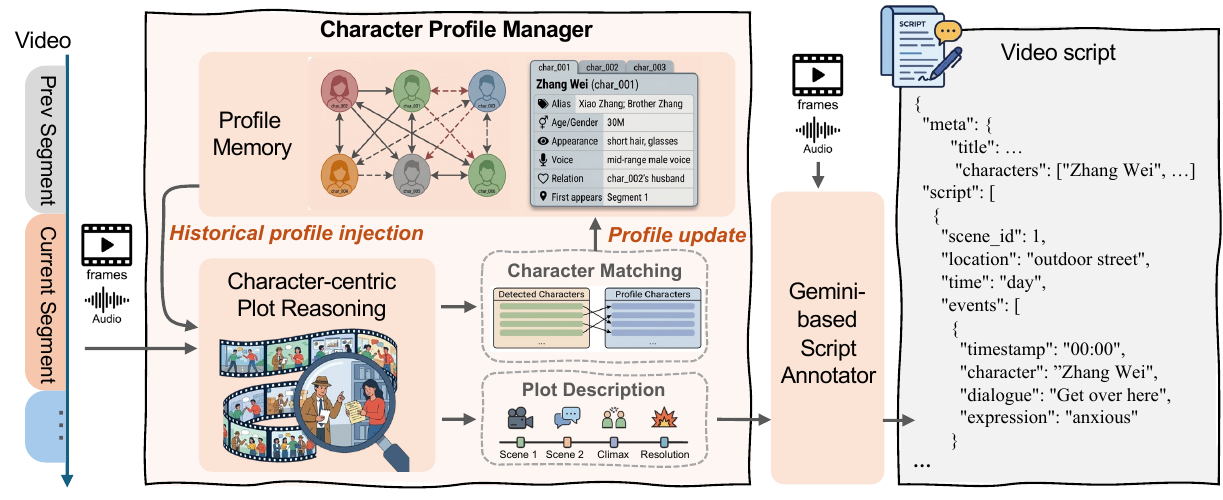}
    \caption{
        Overview of the memory-augmented progressive annotation pipeline. The character profile manager injects historical profiles to guide plot reasoning and dynamically updates character memory. The generated plot description and raw audio-visuals are then fed into a Gemini-based annotator to produce a fine-grained video script.
    }
    \label{fig:method_pipeline}
    \vspace{-5mm}
\end{figure}

\vspace{-2mm}
\subsection{Memory-Augmented Annotation Pipeline}
\label{sec:annotation_pipeline}

Cinematic videos are characterized by intricate narratives, complex character dynamics, and long-range temporal dependencies. Traditional annotation pipelines, which independently process video segments and subsequently merge the textual outputs, often fail to maintain narrative coherence. Furthermore, character re-identification across scenes remains a critical bottleneck, as discriminative cues in movies are inherently multimodal (e.g., speech tone, gait, clothing) rather than purely facial. To overcome these limitations, we propose a novel, memory-augmented progressive annotation pipeline centered around a character profile manager (CPM). This pipeline efficiently processes raw videos into high-quality, chain-of-thought formatted script data. As illustrated in Fig.~\ref{fig:method_pipeline}, our pipeline operates through three progressive stages:

\vspace{-3mm}
\paragraph{Memory-Augmented CPM.}
Given a raw video, we first partition it into semantically cohesive segments (typically <5 minutes) using PySceneDetect based on scene boundaries. From a curated collection of over 10K raw cinematic videos (ranging from minutes to hours), we extract approximately 45K segments. 
For each segment, we employ a short-video expert model (i.e., Gemini-2.5-Pro) to perform character-centric plot reasoning. Crucially, this process is conditioned on the CPM, a memory module that stores cross-segment character information. By integrating current audio-visual inputs with historical profiles retrieved from the CPM, the model accurately resolves character identities and synthesizes a coherent plot description. Simultaneously, the CPM dynamically updates to ensure global consistency by continuously accumulating evolving multimodal attributes (e.g., costume changes) into existing character profiles. Furthermore, it employs a lazy naming strategy for entity resolution, where provisional character IDs are retroactively upgraded to permanent ones and duplicate records are merged upon detecting definitive naming events in the dialogue.

\vspace{-3mm}
\paragraph{Fine-Grained Script Generation.}
With globally consistent plot descriptions established, we proceed to detailed script generation. The generated plot descriptions, which now encapsulate accurate character identities and narrative logic, are fed alongside the original audio-visual content into a powerful MLLM (i.e., Gemini). This step translates the high-level plot into a fine-grained, scene-by-scene script, capturing fine-grained audio-visual cues like character actions, dialogues, and emotional nuances.

\vspace{-3mm}
\paragraph{Thinking Data Construction.}
To endow our target model with robust reasoning capabilities, we construct CoT trajectories driven by plot and character dynamics. We utilize a strong LLM (i.e., DeepSeek) to retroactively distill an "intermediate thinking" process from the generated scripts.
This thinking phase explicitly articulates plot summaries and character relationship mappings. Consequently, we formulate a structured $\text{Video} \rightarrow \text{Thinking} \rightarrow \text{Script}$ CoT dataset. This structured data not only provides high-quality supervision for script generation but also serves as the foundation for our model's reasoning-based training.
\section{The V2S Benchmark: Metrics and Dataset}
\label{sec:benchmark}
Evaluating highly structured, temporally grounded, and semantically rich video scripts poses a unique challenge. Traditional metrics (e.g., BLEU \cite{papineni2002bleu}, ROUGE \cite{lin2004rouge}, or standard temporal action localization metrics) fail to capture the hierarchical dependencies and the open-vocabulary descriptions. To bridge this gap, we introduce a comprehensive Video-to-Script (V2S) benchmarking.

\subsection{Temporally-Aware Hierarchical Evaluation Framework}
\label{subsec:evaluation_metric}

To rigorously and fairly evaluate generated video scripts, we devise a temporally-aware, hierarchical evaluation framework that effectively disentangles semantic fidelity from temporal localization. Conventional metrics relying solely on temporal Intersection-over-Union (tIoU) heavily penalize semantically accurate but slightly temporally shifted predictions, which is sub-optimal for the open-vocabulary and narrative nature of generated scripts. To address this, our pipeline operates through a ``general-to-specific'' four-stage formulation, taking event-level evaluation as our primary instantiation. 

Specifically, the evaluation pipeline encompasses the following four sequential stages:
\begin{itemize}
    \item \textbf{Stage 1: Text-Based Event Matching (The Alignment).} Instead of relying on strict temporal overlap (tIoU) which breaks down under slight narrative delays, we align events based on composite semantic similarity and dynamic programming. This answers ``which predicted event corresponds to which GT event'' while tolerating minor temporal noise.
    \item \textbf{Stage 2: Character Mapping (The Prerequisite).} Generated scripts often utilize open-vocabulary identity descriptions (e.g., predicting ``police officer'' instead of the Ground Truth ``John''). Before any event evaluation can occur, we must establish a unified semantic space for characters to prevent cascading mismatch errors.
    \item \textbf{Stage 3: Field-Level Evaluation (The Semantic Quality).} Once event groups are successfully aligned, we conduct a fine-grained evaluation on the internal fields (character, action, dialogue, expression, audio cue) to strictly assess the semantic correctness of ``what happens''.
    \item \textbf{Stage 4: Temporal Boundary Evaluation (The Localization Quality).} Finally, independent of the semantic details, we evaluate ``when it happens'' by computing the tIoU Hit Rate of the aligned groups, directly penalizing temporal hallucinations and misses.
\end{itemize}

\subsubsection{Stage 1: Text-Content-Guided Event Alignment}
\label{subsubsec:event_alignment}

To establish robust correspondence between predicted and Ground Truth (GT) events while tolerating minor temporal variance, we align events based on composite semantic similarity rather than rigid temporal overlap. This answers ``which predicted event corresponds to which GT event''. For each candidate matching pair (allowing for one-to-many and many-to-one mappings), we compute a composite text score:
\begin{equation}
    \mathcal{S}_{\text{text}} = \frac{\lambda_1 \cdot \mathcal{S}_{\text{dialogue}} + \lambda_2 \cdot \mathcal{S}_{\text{action}}}{\lambda_1 + \lambda_2},
\end{equation}
where $\mathcal{S}_{\text{dialogue}}, \mathcal{S}_{\text{action}} \in [0,1]$ are calculated via normalized Levenshtein distance, and $\lambda_1, \lambda_2$ are empirically set to 5.0 and 3.0, respectively. 

To preserve narrative consistency, we impose a temporal proximity constraint: the absolute temporal distance between any GT and predicted event in a pair must be $\le 30.0$ seconds. To mildly favor temporally proximate matches, we introduce an exponential decay temporal bonus:
\begin{equation}
    \mathcal{S}_{\text{time\_bonus}} = \min\left(0.1, \exp\left(-\frac{\Delta t_{\min}}{15.0}\right) \times 0.1\right).
\end{equation}
The final alignment score is defined as $\mathcal{S}_{\text{align}} = \mathcal{S}_{\text{text}} + \mathcal{S}_{\text{time\_bonus}}$. By formulating a Weighted Interval Scheduling problem, we leverage Dynamic Programming (DP) to efficiently resolve the globally optimal assignment that maximizes the sum of alignment scores, subject to the constraints that neither GT nor predicted indices can overlap, and temporal order must be strictly preserved.

\noindent\textbf{Handling Unaligned Events:} After obtaining the optimal assignment (which forms the \textit{aligned groups}), we explicitly process the remaining unmatched events to ensure comprehensive penalization in subsequent stages. Any GT event that is not matched during the DP process is instantiated as an independent \textit{unaligned GT group}, where its corresponding predicted event is left empty. Conversely, any unmatched predicted event forms an independent \textit{unaligned predicted group}. These unaligned groups are seamlessly integrated into the evaluation pipeline, serving as direct sources for misses (degrading recall) and hallucinations (degrading precision).

\subsubsection{Stage 2: LLM-Assisted Semantic Character Resolution}
\label{subsubsec:character_resolution}

Generated scripts frequently exhibit open-vocabulary identity descriptions (e.g., predicting ``police officer'' instead of the GT ``John''). Before any semantic event evaluation can occur, we must establish a unified semantic space for characters to prevent cascading mismatch errors. We employ a Large Language Model (LLM) to extract all unique character names and their active time intervals, executing two concurrent tasks:

\noindent\textbf{1. Name Categorization:} The LLM parses the extracted entities and categorizes each into one of three types: proper names (e.g., ``John''), singular identity names (e.g., ``officer'', ``man''), and plural identity names (e.g., ``cops'', ``thieves'').

\noindent\textbf{2. Initial Bipartite Mapping Generation:} Concurrently, the LLM constructs an initial bipartite mapping graph $\mathcal{G}$ based purely on semantic equivalence. This safely anchors highly confident, proper name matches (e.g., mapping ``John Smith'' to ``John'') before applying any heuristic rules.

For the remaining unmapped characters, we formulate a fallback similarity score $\mathcal{S}_{\text{fallback}}$ for every possible $(\text{character}_{\text{gt}}, \text{character}_{\text{pred}})$ pair:
\begin{equation}
    \mathcal{S}_{\text{fallback}} = 0.5 \cdot \text{IoU}_{\text{temporal}} + 0.5 \cdot \text{Sim}_{\text{text}},
\end{equation}
where $\text{IoU}_{\text{temporal}}$ is calculated based on the overlapping active intervals, and $\text{Sim}_{\text{text}}$ measures lexical similarity via normalized Levenshtein distance. A greedy matching is then performed based on $\mathcal{S}_{\text{fallback}}$ in descending order. To prevent logical fallacies, this matching is strictly bounded by the prior LLM categorizations:
\begin{enumerate}
    \item Proper names cannot match with other proper names unless explicitly paired in the initial LLM mapping graph $\mathcal{G}$.
    \item Proper names cannot match with plural identity names.
    \item Singular identity names cannot match with plural identity names.
\end{enumerate}
Fallback matches are only accepted if $\mathcal{S}_{\text{fallback}}$ exceeds a threshold of $\tau = 0.05$. 

\noindent\textbf{Global Mapping Dictionary:} The definitive output of this stage is a global character mapping dictionary that projects predicted characters onto the GT semantic space (e.g., \texttt{\{"pred\_cop": "gt\_police\_officer", "man in black": "John", "thieves": "robbers"\}}). This dictionary is rigidly applied in Stage 3 to ensure consistent tracking.

\subsubsection{Stage 3: Multi-dimensional Field Evaluation}
\label{subsubsec:field_evaluation}

Because the event alignment permits one-to-many and many-to-one mappings, a single aligned event group may contain multiple fragmented events. Before evaluating specific fields, we merge the internal items by grouping them according to characters. For textual fields, descriptions sharing the same character are concatenated into a single continuous string, while events involving different characters are appended as separate elements in a list. The unified temporal boundary of the merged group is calculated by taking the minimum start time and maximum end time across all its constituent events.

To intuitively conceptualize this consolidation, consider the following 1-to-N alignment scenario:

\vspace{1mm}
\noindent\textit{Before Merging:}
\begin{itemize}
    \item \textbf{Ground Truth (1 Event):}
    \begin{itemize}
        \item Event 1: Time \texttt{[00:05 - 00:15]} | Character: Officers | Action: ``secure the perimeter and enter the building''
    \end{itemize}
    \item \textbf{Prediction (2 Events):}
    \begin{itemize}
        \item Event 1: Time \texttt{[00:04 - 00:08]} | Character: Officer A | Action: ``secures the perimeter''
        \item Event 2: Time \texttt{[00:09 - 00:14]} | Character: Officer B | Action: ``enters the building''
    \end{itemize}
\end{itemize}
\vspace{1mm}

During the merging phase, the temporal boundaries are expanded ($\min(00:04, 00:09)$ to $\max(00:08, 00:14)$). Because the predicted characters (Officer A and Officer B) are different, their actions are appended into a list of two separate items, whereas the GT remains a single item.

\vspace{1mm}
\noindent\textit{After Merging:}
\begin{itemize}
    \item \textbf{Merged Ground Truth:} Time \texttt{[00:05 - 00:15]} | Action List: \texttt{["secure the perimeter and enter the building"]}
    \item \textbf{Merged Prediction:} Time \texttt{[00:04 - 00:14]} | Action List: \texttt{["secures the perimeter", "enters the building"]}
\end{itemize}
\vspace{1mm}

\noindent\textbf{Field Evaluation Formats:} After merging, we evaluate five semantic fields:
\begin{itemize}
    \item \textbf{Character:} Exact string matching (after applying the global mapping dictionary).
    \item \textbf{Dialogue:} Normalized Levenshtein edit distance.
    \item \textbf{Action, Expression, Audio Cue:} We employ an LLM to assess the semantic similarity $\mathcal{S} \in [0, 1]$ between the merged GT and predicted text lists.
\end{itemize}

\noindent\textbf{Similarity Score Calculation (1-to-N Example):} Revisiting the action field example above, we compare 1 GT action against 2 predicted actions. The evaluator computes a similarity matrix for all valid pairs. Suppose the semantic similarities are evaluated as $\mathcal{S}(\text{GT}_1, \text{Pred}_1) = 0.90$ and $\mathcal{S}(\text{GT}_1, \text{Pred}_2) = 0.85$. To prevent unbounded scores and strictly penalize redundant predictions, our evaluation utilizes a greedy matching strategy. The single GT item is exclusively matched to the predicted item with the highest similarity score. Thus, $\text{GT}_1$ is matched with $\text{Pred}_1$, leaving $\text{Pred}_2$ unmatched. The total matched score $\mathcal{S}_{\text{group}}$ for this field within the event group is simply the maximum similarity from the matched pair: $\mathcal{S}_{\text{group}} = \max(0.90, 0.85) = 0.90$.

\noindent\textbf{Precision, Recall, and F1 Formulation:} For a specific field, let $\mathcal{S}_{\text{total}}$ be the cumulative sum of $\mathcal{S}_{\text{group}}$ across all aligned event groups. We calculate global precision and recall utilizing total item counts:
\begin{itemize}
    \item \textbf{Precision Denominator ($N_{\text{pred\_total}}$):} The sum of $N_{\text{pred}}$ from all \textit{aligned} predicted event groups. As demonstrated, any unmatched redundant items within an aligned group inflate this denominator without contributing to $\mathcal{S}_{\text{total}}$, naturally penalizing hallucinations.
    \item \textbf{Recall Denominator ($N_{\text{gt\_total}}$):} The sum of $N_{\text{gt}}$ across the \textit{entire original Ground Truth script}. Crucially, this sum includes both the aligned GT events and the completely unaligned (missed) GT events, rigorously penalizing omissions.
\end{itemize}
The single-video metrics are calculated as:
\begin{equation}
    \text{Precision} = \frac{\mathcal{S}_{\text{total}}}{N_{\text{pred\_total}}}, \quad \text{Recall} = \frac{\mathcal{S}_{\text{total}}}{N_{\text{gt\_total}}}, \quad F1_{\text{field}} = \frac{2 \cdot \text{Precision} \cdot \text{Recall}}{\text{Precision} + \text{Recall}}.
\end{equation}

For global metrics across the entire dataset, we aggregate the numerators and denominators across all video samples \textit{before} performing the final division.

\subsubsection{Stage 4: Temporal Boundary Evaluation (tIoU Hit Rate)}
\label{subsubsec:temporal_evaluation}

Independent of semantic details, this stage measures group-level temporal localization (``when it happens''). For a given strictness threshold $t$, an event group is considered a ``Hit'' if the temporal IoU of its GT and predicted events satisfies $\text{tIoU} = \frac{\text{Intersection}}{\text{Union}} \ge t$. We compute the temporal metrics as:
\begin{equation}
    P_{\text{time}}@t = \frac{\text{Hit}_{\text{pred}}(t)}{|\text{Total Predicted Groups}|}, \quad R_{\text{time}}@t = \frac{\text{Hit}_{\text{gt}}(t)}{|\text{Total GT Groups}|}.
\end{equation}

Note that ``Total GT Groups'' includes both matched groups and unaligned GT groups (misses), naturally degrading recall if events are not detected. Similarly, unaligned predicted groups (hallucinations) degrade precision. The final tIoU Hit Rate is calculated as:
\begin{equation}
    tIoU@t = \frac{2 \cdot P_{\text{time}}@t \cdot R_{\text{time}}@t}{P_{\text{time}}@t + R_{\text{time}}@t}.
\end{equation}

\subsection{Construction and Quality Assurance}

To systematically evaluate models, we construct a meticulously curated benchmark of 10 full-length cinematic works (19.9 hours total) covering diverse genres (anime, action, suspense, drama). Unlike traditional flat video captioning datasets, our benchmark introduces a dense, hierarchical structure. It decomposes into 1.4k distinct scenes and over 16.8k structural events, averaging an exceptionally high density of 14.1 events per minute. Crucially, it incorporates nuanced modalities rarely present in existing benchmarks, such as facial expressions, audio cues, and subtexts. To facilitate a comprehensive evaluation across varying temporal horizons, we systematically segment and sample the annotated works. Ultimately, our evaluation benchmark comprises a multi-granularity test bed: 200 5-minute, 100 10-minute, 50 15-minute, 40 20-minute, 30 25-minute, and 25 30-minute cinematic clips, specifically designed to stress-test the long-context robustness of multimodal models.

Annotating such dense, multi-modal scripts is notoriously labor-intensive. We adopt an efficient model-in-the-loop paradigm where our automated data engine generates initial dense pseudo-labels, which are subsequently refined by trained annotators. To guarantee utmost reliability, we eschew standard automated consistency checks in favor of a rigorous expert-in-the-loop verification mechanism. Every script undergoes a final comprehensive review by a senior domain expert to cross-reference temporal boundaries and ensure long-term semantic coherence, yielding a benchmark of unprecedented quality. 

\section{OmniScript}
\label{sec:method}
\subsection{Architecture}
\label{sec:architecture}
\begin{figure}[t!]
    \centering
    \vspace{0mm}  
    \includegraphics[width=1.0\textwidth]{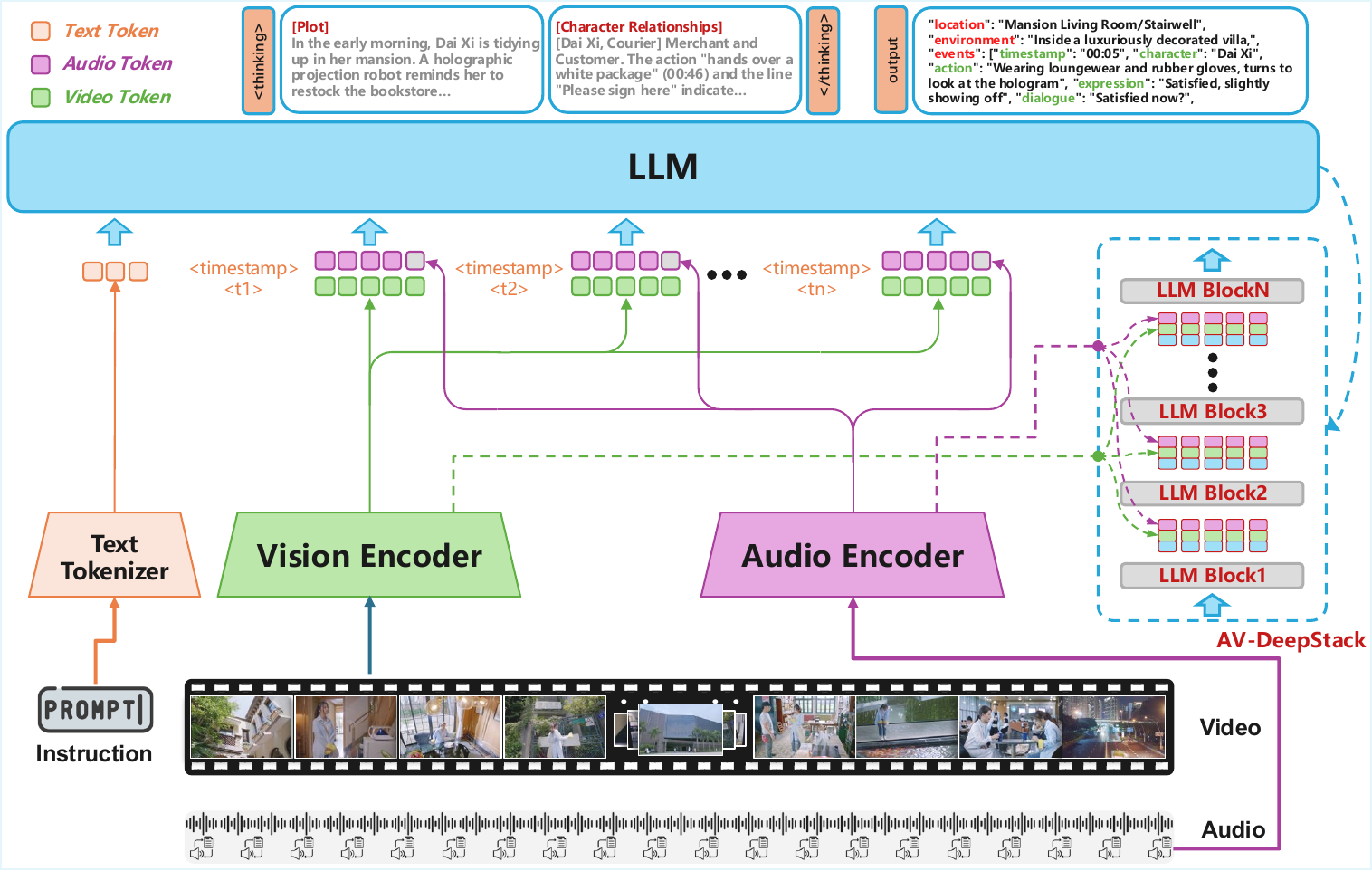}
    \caption{Overview of the proposed architecture. Instruction, video, and audio are encoded into multimodal tokens and fused in the LLM via AV-DeepStack across multiple layers. The model first performs multimodal plot and character-relationship reasoning and then generates structured script outputs, including location, environment, events, character, action, expression, and dialogue for each event.}
  \label{fig:architecture} 
  \vspace{-6mm}
\end{figure}

As shown in Fig.~\ref{fig:architecture}, OmniScript follows a unified multimodal reasoning-to-generation pipeline for long-form cinematic video understanding. Given a video clip, the system first encodes visual frames and raw audio into temporally indexed token sequences, then injects these tokens into a large language model (LLM) to perform joint narrative reasoning. The model output is organized into a structured script that explicitly contains scene context (e.g., location and environment) and event-level elements (character, action, expression, dialogue, and audio-aware cues), matching the target schema in Section~\ref{sec:script_model}.

\noindent\textbf{Multimodal Temporal Alignment.}
Beyond conventional vision-only pipelines, we introduce an audio pathway and enforce strict timestamp-level alignment between video and audio features. 
Specifically, we utilize a pre-trained Whisper~\cite{radford2023robust} encoder to extract audio information.
For each temporal unit, the encoder constructs a paired representation $(v_t, a_t)$, where $v_t$ and $a_t$ denote visual and audio embeddings at the same timestamp. This one-to-one alignment preserves cross-modal synchrony for dialogues, off-screen narration, environmental sounds, and background music, which are all critical for script-level narrative grounding.

\noindent\textbf{AV-DeepStack Injection.}
Our model is built on the Qwen3-VL~\cite{bai2025qwen3}, a vision-language model with native visual-textual reasoning, and adopts a DeepStack-style fusion strategy that injects visual features into multiple LLM layers rather than only at the input stage.
Building on this strong VL prior, we extend the architecture to full audio-visual-language modalities. 
Specifically, after temporal alignment, audio tokens are paired with visual tokens and jointly injected across stacked transformer layers via our AV-DeepStack module. 
In each layer, the language stream is conditioned by both modalities through residual multimodal adapters, enabling repeated cross-modal interaction during deep semantic inference. This extension preserves the long-context reasoning strengths of the original DeepStack design while adding explicit auditory perception, which is essential for dialogue understanding (including off-screen speech/narration), acoustic event grounding, and BGM-driven emotion interpretation.

\noindent\textbf{Reasoning-Guided Structured Decoding.}
To maintain global-local consistency, the decoder employs a Chain-of-Thought (CoT) paradigm. Rather than directly predicting event fields, the model first generates an intermediate reasoning trace conditioned on the video segment, comprising (i) a plot progression summary and (ii) an explicit character-relationship state. This trace acts as a structural scaffold for the subsequent coarse-to-fine generation of the temporally grounded script. Specifically, the model establishes the scene context before predicting ordered event records with structured fields (character, action, expression, dialogue, and audio cues). This formulation aligns storyline evolution with event-level details, mitigates long-context ambiguity, and enhances robustness in complex cinematic scenarios involving implicit speaker turns, off-screen audio, and dynamic interpersonal relations.

\subsection{Progressive Training and Alignment}

OmniScript is optimized via a progressive four-stage pipeline followed by reinforcement learning refinement:

\noindent \textbf{1. Modality Alignment}: 
To integrate the additional audio modality into the vision-language backbone, we first conduct a modality alignment stage. 
Using approximately 1M bilingual (CN/EN) in-domain cinematic samples with timestamped ASR supervision, we train only the newly introduced audio modules (audio projector) while freezing the original components (Whisper encoder, ViT, and LLM). This parameter-efficient setup preserves pretrained visual-language reasoning while establishing stable cross-modal correspondences. To prevent over-reliance on visual evidence, we randomly mask video frames, encouraging the model to exploit complementary audio cues under partial visual observations.

\noindent \textbf{2. Multimodal Pretraining}: 
Following alignment, we perform large-scale multimodal pretraining to enhance audio-visual-language understanding in long-form cinematic videos. This stage addresses three objectives: (i) unifying cross-modal semantics across diverse narrative styles, (ii) enhancing temporal grounding for event-level localization, and (iii) improving character-centric action and dialogue comprehension. We curate a bilingual corpus of $2.4\text{M}$ in-domain videos and optimize the model using a multi-task objective encompassing ASR (with and without timestamps), video summarization, dense video captioning, and temporal grounding. Unlike the alignment stage, we fully fine-tune the core components to deeply adapt representations and reasoning layers to the cinematic domain. Random frame masking is retained as a regularization technique.

\noindent \textbf{3. Supervised Fine-Tuning (SFT)}: 
We subsequently apply supervised fine-tuning (SFT) to adapt the pretrained model for structured script generation. To improve schema adherence, narrative coherence, and audio-aware descriptions (e.g., BGM and sound effects), we formulate SFT as a Chain-of-Thought (CoT) process. Empirically, plot-conditioned generation yields superior results; thus, we train the fully fine-tuned LLM to first generate an intermediate reasoning trace (capturing plot progression and character relations) before decoding the final structured fields. 
The SFT dataset comprises $45\text{k}$ in-domain videos ($21\text{k}$ horizontal movie/TV clips and $24\text{k}$ vertical short dramas), covering both long cinematic narratives and fast-paced short-video storytelling.
The videos are annotated following the pipeline introduced in Section~\ref{sec:annotation_pipeline}. 
Furthermore, we introduce random subtitle masking to reduce reliance on explicit textual cues and improve robustness against missing or noisy subtitles.

\noindent \textbf{4. Reinforcement Learning (RL)}: 
To improve the fine-grained descriptive capabilities of the generated scripts, we apply reinforcement learning with verifiable rewards during post-training. Specifically, we optimize the model using GRPO~\cite{shao2024deepseekmath} on a small, high-quality dataset of human-annotated scripts. A primary bottleneck in long-sequence, open-ended generation is the design of an effective reward function. Existing methods relying on global semantic similarity often bias toward dominant features, thereby masking subtle errors related to short-duration events. To overcome this, we propose a temporally segmented reward mechanism that evaluates key components across the entire video timeline. The reward is computed using the Multi-dimensional Field Evaluation score (Section \ref{subsec:evaluation_metric}), which performs event-level, one-to-one matching between the generated and ground-truth scripts. This localized alignment allows for the rigorous penalization of fine-grained errors in both recall and precision.

\subsection{Long Video Processing}

\begin{figure}[t]
    \centering
    \begin{subfigure}[t]{0.49\textwidth}
        \centering
        \includegraphics[width=\linewidth]{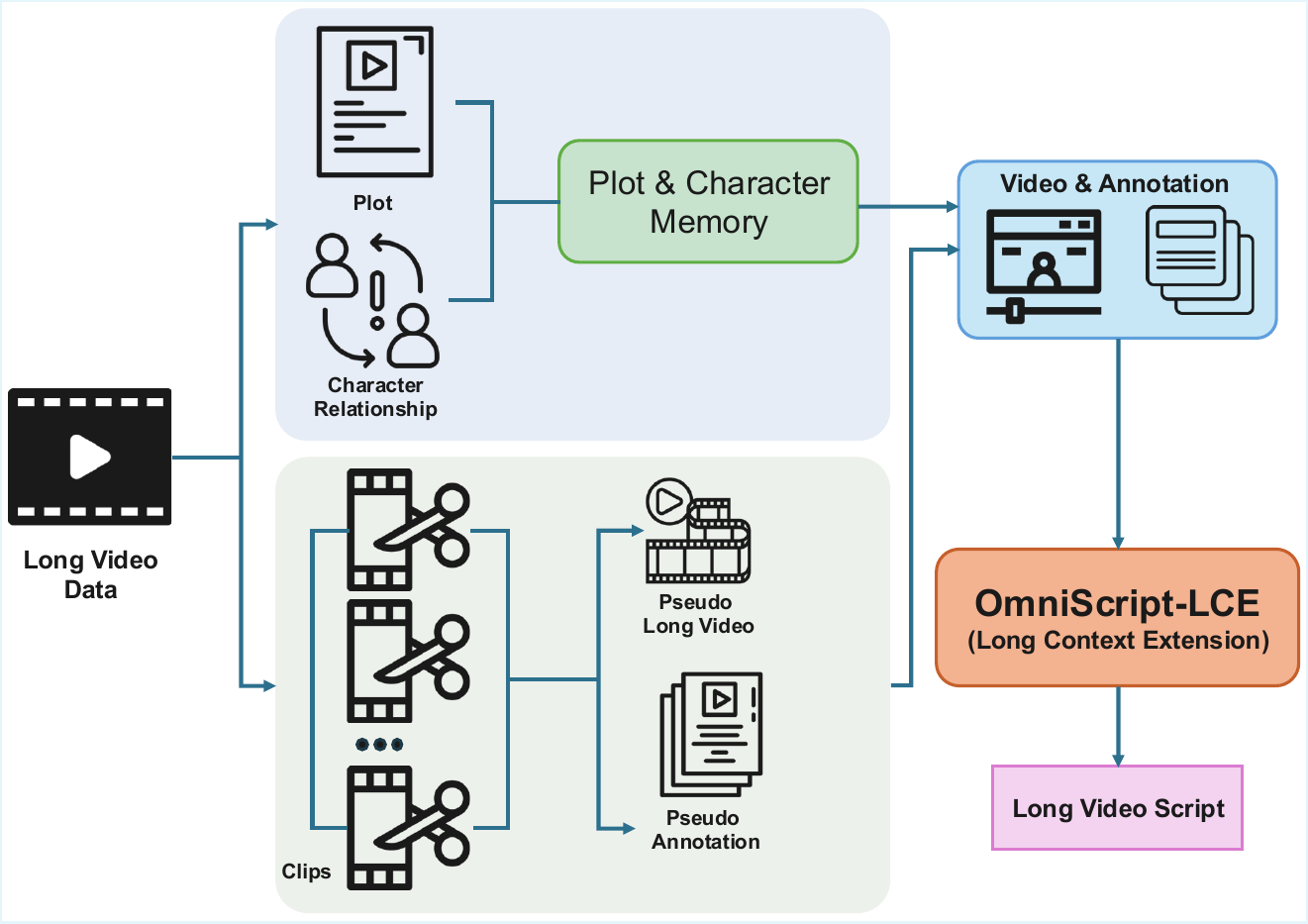}
        \caption{Strategy 1: Long context extension}
    \end{subfigure}
    \hfill
    \begin{subfigure}[t]{0.49\textwidth}
        \centering
        \includegraphics[width=\linewidth]{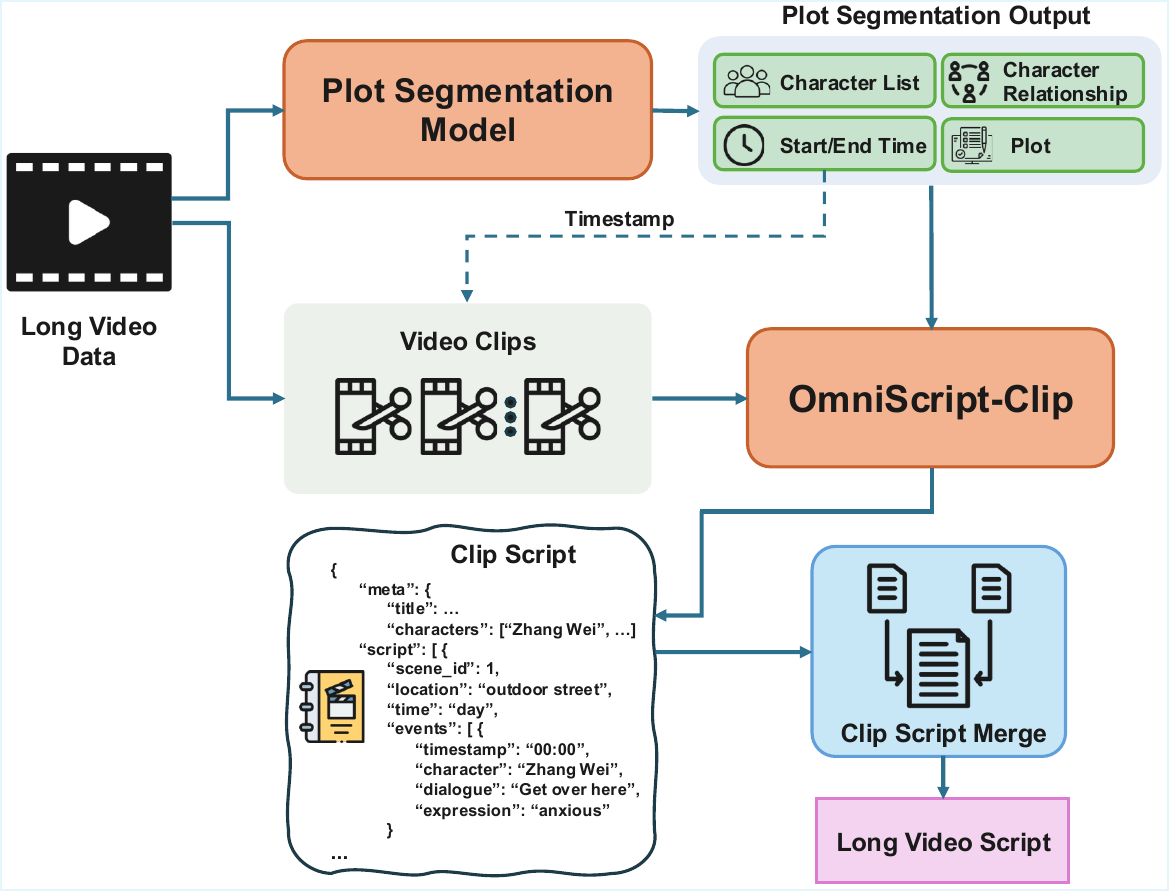}
        \caption{Strategy 2: Two-stage script generation}
    \end{subfigure}
    \caption{Comparison of two strategies for extending OmniScript to long videos. \textbf{Left (Strategy 1):} Direct context extension, trained with long-video annotations (including memory-refine labels) and cross-video composition to create pseudo videos and pseudo scripts. \textbf{Right (Strategy 2):} Two-stage inference: first train a plot-segmentation model to predict each segment's timestamps, plot, character list, and relations, then feed each clip with its plot/character information into OmniScript for segment-level generation, and finally merge all segment outputs.}
    \label{fig:longvideo_extension}
\end{figure}

With forementioned design, our omniscript can process video whose duration is less than 5 minute.
To extend OmniScript to longer videos, we investigate two practical strategies shown in Figure~\ref{fig:longvideo_extension}.

\noindent\textbf{Strategy 1: Long-context extension (OmniScript-LCE).}
We directly scale the input context window and train the model with long-video supervision.
Specifically, we collect long-form annotations that include (i) global storyline descriptions, (ii) segment-level plot transitions, and (iii) memory-refine labels for correcting historical inconsistencies over long horizons.
Because fully annotated long videos are limited, we further construct pseudo long videos by cross-video composition: multiple short clips with coherent themes are concatenated, and pseudo scripts are generated by merging their aligned plot/character annotations.
This strategy keeps a single-stage generation pipeline, but requires stronger long-range reasoning and larger computation during training and inference.

\noindent\textbf{Strategy 2: Two-stage script generation (OmniScript-TSG).}
We decompose long-video generation into planning and writing.
In stage~1, a plot-segmentation model predicts a sequence of segments with timestamps, segment plots, active characters, and inter-character relations.
In stage~2, each segment is processed independently by OmniScript, conditioned on both visual content and stage-1 structural prompts, to produce segment-level scripts.
Finally, we merge all segment outputs with a lightweight post-processing module that enforces temporal consistency (e.g., names, coreference, and event ordering) and produces a unified long-form script.
This decomposition reduces context-length pressure and improves controllability over long narratives.

For fair comparison, both strategies share the same base OmniScript backbone and training objectives for script generation.
The key difference is whether long-range dependency is handled implicitly in a single pass (Strategy~1) or explicitly through segmentation and composition (Strategy~2).
\section{Experiments}
\label{sec:experiment}

\begin{table}[tbp]
  \centering
  \vspace{-1em}
  \caption{Comparison of Event-level metrics for SOTA models on 5-minute videos. Omni indicates whether a model supports audio input ($\checkmark$: yes, $\times$: no). In model names, -T denotes Thinking mode. For MoE models, parameters are reported as total parameters/activated parameters. $\dag$ indicates that a 5-minute video input to the model is first divided into 1-minute segments, which are then sequentially fed into the model. The outputs of all segments are then concatenated to form the final output.}
  \vspace{-1em}
  \label{tab:sota_event_level}
  \resizebox{\linewidth}{!}{
  \begin{tabular}{lccccccccc}
    \toprule
    \textbf{Model} & \textbf{Param./B} & \textbf{Omni} & \textbf{Char.} & \textbf{Dia.} & \textbf{Act.} & \textbf{Exp.} & \textbf{Aud.} & \textbf{Overall} & \textbf{tIoU@0.1} \\
    \midrule
    \multicolumn{10}{l}{\textit{Proprietary Models}} \\
    Gemini-3-flash~\cite{gemini3flash}   &  & $\checkmark$ & 28.8 & 50.3 & 28.2 & 25.5 & 11.2 & 28.8 & 44.3 \\
    Gemini-3-pro~\cite{gemini3pro}     &  & $\checkmark$ & 39.8 & 68.8 & 37.4 & 35.4 & 13.3 & 38.9 & 64.4 \\
    Gemini-2.5-flash~\cite{gemini25flash} &  & $\checkmark$ & 40.1 & \textbf{75.5} & \underline{42.8} & \underline{36.5} & \textbf{22.8} & \textbf{43.6} & \textbf{74.3} \\
    Gemini-2.5-pro~\cite{gemini25pro}  &  & $\checkmark$ & \underline{41.7} & \underline{75.0} & 41.9 & \textbf{39.0} & \underline{17.0} & \underline{42.9} & \underline{73.4} \\
    Seed-1.8~\cite{seed1-8} &  & $\times$ & 40.9 & 54.4 & 35.1 & 29.6 & 12.4 & 34.5 & 50.7 \\
    Seed-2.0-pro~\cite{seed2pro} &  & $\times$ & \textbf{47.4} & 68.1 & \textbf{42.9} & 35.7 & 10.3 & 40.9 & 67.1 \\
    \midrule
    \multicolumn{10}{l}{\textit{Open-source Models}} \\
    \textcolor{gray}{MiniCPM-O-4.5$^\dag$ \cite{yu2025minicpm}} & \textcolor{gray}{9} & \textcolor{gray}{$\checkmark$} & \textcolor{gray}{26.7} & \textcolor{gray}{59.1} & \textcolor{gray}{32.3} & \textcolor{gray}{23.4} & \textcolor{gray}{17.7} & \textcolor{gray}{31.8} & \textcolor{gray}{54.7} \\
\textcolor{gray}{TimeChat-Captioner$^\dag$ \cite{yao2026timechat}} & \textcolor{gray}{8} & \textcolor{gray}{$\checkmark$} & \textcolor{gray}{31.2} & \textcolor{gray}{36.8} & \textcolor{gray}{36.2} & \textcolor{gray}{31.8} & \textcolor{gray}{25.4} & \textcolor{gray}{32.3} & \textcolor{gray}{64.0} \\
    Qwen3-Omni-T~\cite{xu2025qwen3} & 30/3 & $\checkmark$ & 3.2 & 3.5 & 3.8 & 6.8 & 2.3 & 3.9 & 3.0 \\
    Qwen3-Omni~\cite{xu2025qwen3} & 30/3 & $\checkmark$ & 4.9 &	3.4 &	5.5	& 7.3	& 4.4	& 5.1	& 12.8 \\
    MiniCPM-O-4.5~\cite{yu2025minicpm} & 9 & $\checkmark$ & 3.1 & 8.0 & 2.9 & 3.2 & 2.4 & 3.9 & 3.2 \\
    TimeChat-Captioner~\cite{yao2026timechat} & 8 & $\checkmark$ & 6.9 & 6.9 & 6.6 & 9.7 & \underline{8.2} & 7.7 & 16.1 \\
    Qwen3VL~\cite{bai2025qwen3} & 8        & $\times$ & 30.4 & 49.6 & 26.9 & 25.3 & 6.6 & 27.7 & 47.6 \\
    Qwen3VL-T~\cite{bai2025qwen3} & 32 & $\times$ & 24.4 & 37.5 & 22.1 & 18.9 & 7.0 & 22.0 & 34.6 \\
    Qwen3VL~\cite{bai2025qwen3} & 32       & $\times$ & 37.1 & 57.1 & 31.3 & 28.7 & 7.2 & 32.3 & 52.5 \\
    Qwen3VL-T~\cite{bai2025qwen3}        & 235/22 & $\times$ & 35.7 & 57.6 & 27.4 & 23.9 & 6.5 & 30.2 & 54.8 \\
    Qwen3VL~\cite{bai2025qwen3}          & 235/22 & $\times$ & \underline{38.1} & \underline{58.6} & \underline{33.0} & \underline{29.1} & 6.0 & \underline{33.0} & \underline{62.0} \\
    \textbf{Ours} & 8 & $\checkmark$ & \textbf{39.2} & \textbf{72.2} & \textbf{33.7} & \textbf{31.9} & \textbf{11.6} & \textbf{37.7} & \textbf{69.3} \\
    \bottomrule
  \end{tabular}
  }
  \vspace{-2em}
\end{table}

\subsection{Implementation Details}
We initialize the model from Qwen3VL-8B and initialize the audio encoder from Whisper large-v3. Our training pipeline follows the three-stage recipe detailed in Sec.~\ref{sec:method}. In modality alignment, we freeze the LLM backbone and optimize modality projectors to align visual/audio representations with text space, while applying random frame masking to improve robustness under partial observations. In pretraining, we perform full fine-tuning on approximately 2.4M bilingual (Chinese/English) in-domain videos with a unified multi-task objective, including ASR (with/without timestamps), dense captioning, summarization, and temporal grounding. In SFT, we further train on about 45K curated videos (21K movie/TV clips and 24K short-form drama clips) using schema-level supervision with CoT-style intermediate traces (plot evolution and character relation reasoning) before final structured decoding. All performance results and comparisons in this section are conducted on our proposed benchmark.

\begin{table}[tbp]
  \centering
  \caption{Comparison of Scene-level metrics for SOTA models on 5-minute videos. Omni indicates whether a model supports audio input ($\checkmark$: yes, $\times$: no). $\dag$ indicates that a 5-minute video input to the model is first divided into 1-minute segments, which are then sequentially fed into the model. The outputs of all segments are then concatenated to form the final output.}
  \vspace{-1em}
  \label{tab:sota_scene_level}
  \resizebox{\linewidth}{!}{
  \begin{tabular}{lccccccccc}
    \toprule
    \textbf{Model} & \textbf{Param./B} & \textbf{Omni} & \textbf{Loc.} & \textbf{Type} & \textbf{Env.} & \textbf{Time} & \textbf{Mood} & \textbf{Overall} & \textbf{tIoU@0.1} \\
    \midrule
    \multicolumn{10}{l}{\textit{Proprietary Models}} \\
    Gemini-3-flash~\cite{gemini3flash} &  & $\checkmark$ & 54.6 & 59.8 & 42.7 & 54.9 & 50.4 & 52.5 & 70.3 \\
    Gemini-3-pro~\cite{gemini3pro} &  & $\checkmark$ & \textbf{58.8} & \textbf{63.1} & 46.9 & \underline{61.6} & \textbf{54.8} & \underline{57.0} & \underline{75.3} \\
    Gemini-2.5-flash~\cite{gemini25flash} &  & $\checkmark$ & 52.8 & 57.1 & 45.7 & 56.1 & 50.3 & 52.3 & 69.6 \\
    Gemini-2.5-pro~\cite{gemini25pro} &  & $\checkmark$ & 56.6 & \underline{62.4} & \textbf{50.8} & 60.1 & \underline{54.6} & 56.9 & 74.1 \\
    Seed-1.8~\cite{seed1-8} &  & $\times$ & \underline{57.9} & 58.6 & 47.7 & 58.7 & 52.8 & 55.1 & 74.0 \\
    Seed-2.0-pro~\cite{seed2pro} &  & $\times$ & 57.7 & 62.2 & \underline{49.2} & \textbf{62.7} & 54.3 & \textbf{57.2} & \textbf{75.5} \\
    \midrule
    \multicolumn{10}{l}{\textit{Open-source Models}} \\
    \textcolor{black!40}{MiniCPM-O-4.5$^\dag$\cite{yu2025minicpm}} & \textcolor{black!40}{9} & \textcolor{black!40}{$\checkmark$} & \textcolor{black!40}{39.9} & \textcolor{black!40}{51.7} & \textcolor{black!40}{34.7} & \textcolor{black!40}{45.2} & \textcolor{black!40}{43.8} & \textcolor{black!40}{43.1} & \textcolor{black!40}{63.4} \\
        \textcolor{black!40}{TimeChat-Captioner$^\dag$\cite{yao2026timechat}} & \textcolor{black!40}{8} & \textcolor{black!40}{$\checkmark$} & \textcolor{black!40}{47.3} & \textcolor{black!40}{55.5} & \textcolor{black!40}{41.8} & \textcolor{black!40}{47.1} & \textcolor{black!40}{48.4} & \textcolor{black!40}{48.0} & \textcolor{black!40}{69.6} \\
    Qwen3-Omni~\cite{xu2025qwen3} & 30/3   & $\checkmark$ & 18.4	& 26.0	& 14.6	& 23.6	& 22.4	& 21.0	& 29.6 \\
    MiniCPM-O-4.5~\cite{yu2025minicpm} & 9 & $\checkmark$ & 10.3 & 22.0 & 8.4 & 17.7 & 17.4 & 15.1 & 32.0 \\
    TimeChat-Captioner~\cite{yao2026timechat} & 8 & $\checkmark$ & 19.9 & 29.5 & 17.3 & 28.8 & 30.5 & 25.2 & 46.6 \\
    Qwen3VL~\cite{bai2025qwen3} & 8      & $\times$ & 41.3 & 49.7 & 31.8 & 39.8 & 41.7 & 40.9 & 60.6 \\
    Qwen3VL~\cite{bai2025qwen3} & 32     & $\times$ & 50.4 & \underline{58.7} & \underline{42.7} & 55.4 & 47.9 & 51.0 & 71.1 \\
    Qwen3VL~\cite{bai2025qwen3} & 235/22 & $\times$ & \underline{52.6} & \textbf{60.2} & \textbf{45.4} & \underline{57.9} & \textbf{50.9} & \textbf{53.4} & \underline{72.8} \\
    \textbf{Ours} & 8 & $\checkmark$ & \textbf{54.0} & 58.4 & 41.9 & \textbf{58.1} & \underline{49.5} & \underline{52.4} & \textbf{74.6} \\
    \bottomrule
  \end{tabular}
  }
  \vspace{-1em}
\end{table}

\subsection{Performance comparison on 5-Minute Videos}
Tables~\ref{tab:sota_event_level} and~\ref{tab:sota_scene_level} compare our model with representative proprietary and open-source models under a unified protocol. We report F1 scores for all event/scene fields, and \textit{Overall} is their mean, reflecting the accuracy of event content understanding. For temporal localization, we use tIoU@0.1. A key takeaway is parameter efficiency: our model uses only 8B parameters, yet delivers strong performance on both event content quality and temporal localization.

\noindent \textbf{Event-level comparison.} In Table~\ref{tab:sota_event_level}, our 8B model achieves 37.0 Overall score and 69.0 tIoU@0.1. It substantially outperforms open-source models of much larger scales, gaining +4.0 in Overall score and +7.0 in tIoU@0.1 compared to Qwen3VL-235B-A22B. Against proprietary models, it achieves stronger dialogue understanding than Gemini-3-pro \cite{gemini3pro} and Seed-2.0-pro \cite{seed2pro}, and also surpasses them in temporal localization.
MiniCPM-O-4.5 and TimeChat-Captioner experience drastic performance drops due to their limited capabilities in processing extended video sequences, achieving Overall event-level metrics of merely 3.9 and 7.7, respectively. Their temporal localization is similarly impaired, with Event tIoU@0.1 plummeting to 3.2 and 16.1. 
To further probe the understanding capabilities of MiniCPM-O-4.5 and TimeChat-Captioner while mitigating their long-context limitations, we also report their performance under a segmented protocol (denoted with $\dag$), where the 5-minute video is artificially divided into 1-minute non-overlapping chunks and fed sequentially. While this segmented approach circumvents their context window constraints and significantly boosts their performance (e.g., TimeChat-Captioner$^\dag$ improves to an Overall event score of 32.3), our model—processing the continuous 5-minute video natively—still consistently outperforms them across all narrative fields. 
We also observe that thinking variants are often worse than their non-thinking counterparts, and existing omni-input variants (e.g., Qwen3-Omni) are notably weaker than similarly sized non-omni models.

\noindent \textbf{Scene-level comparison.} In Table~\ref{tab:sota_scene_level}, our model reaches a 52.6 Overall score and 73.6 tIoU@0.1. Relative to Qwen3VL-235/22B, it shows better temporal boundary quality (+0.8 on tIoU@0.1) at a much smaller scale. Moreover, our holistic approach yields substantially higher spatial-temporal coherence than TimeChat-Captioner$^\dag$ (i.e., Scene tIoU@0.1 of 74.6 vs. 69.6), showing that our method is uniquely equipped to capture the global narrative continuity that is inevitably severed when artificially dividing videos into isolated chunks. It is worth noting that scene attributes requiring subtle visual-audio context integration, such as environment and mood, remain challenging across all evaluated models.

\begin{figure*}[!h]
    \centering
    \begin{subfigure}[t]{0.48\textwidth}
        \centering
        \includegraphics[width=\linewidth]{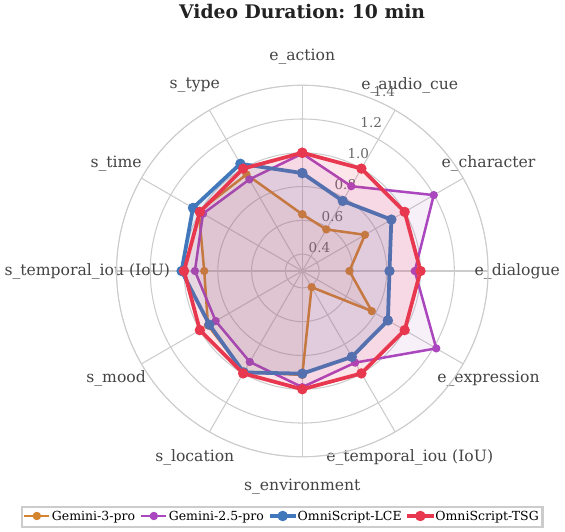}
        \caption{10 min}
    \end{subfigure}
    \hfill
    \begin{subfigure}[t]{0.48\textwidth}
        \centering
        \includegraphics[width=\linewidth]{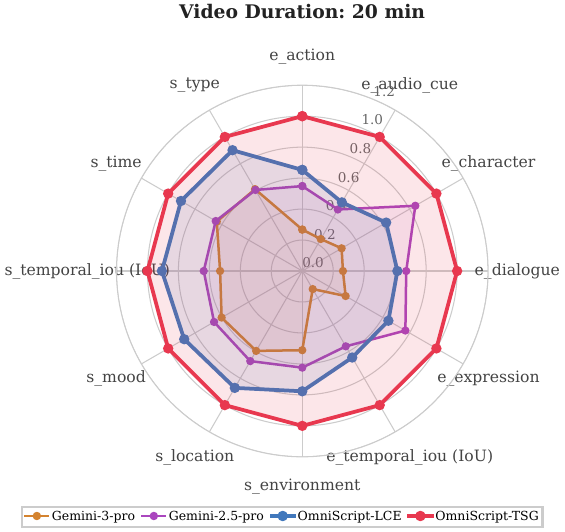}
        \caption{20 min}
    \end{subfigure}
    
    \vspace{0.3cm} 
    
    \begin{subfigure}[t]{0.48\textwidth}
        \centering
        \includegraphics[width=\linewidth]{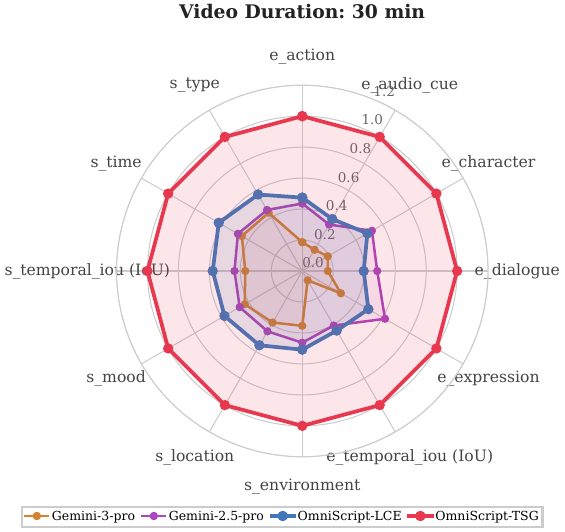}
        \caption{30 min}
    \end{subfigure}
    \hfill
    \begin{subfigure}[t]{0.48\textwidth}
        \centering
        \includegraphics[width=\linewidth]{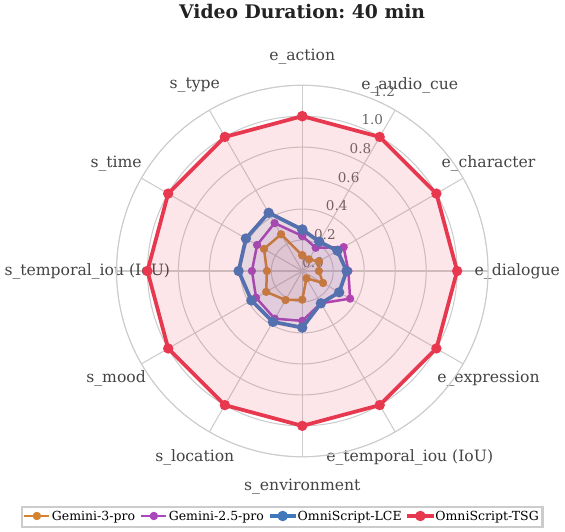}
        \caption{40 min}
    \end{subfigure}
    
    \caption{Performance comparison across video durations on multiple metric dimensions.}
    \label{fig:longvideo_radar}
\end{figure*}

\subsection{Performance Comparison on Longer Videos}
\label{subsec:longer_videos}

\begin{figure*}[t]
    \centering
    \begin{subfigure}[t]{\textwidth}
        \centering
        \includegraphics[width=\linewidth]{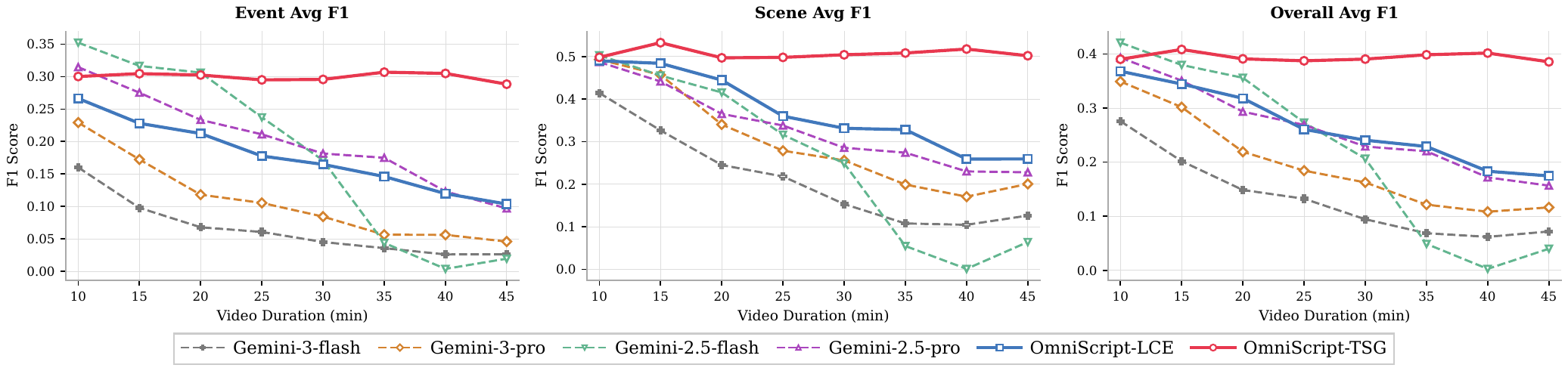}
        \caption{Event-level, scene-level, and overall average F1 scores across long-video durations.}
    \end{subfigure}

    \vspace{0.6em}

    \begin{subfigure}[t]{\textwidth}
        \centering
        \includegraphics[width=\linewidth]{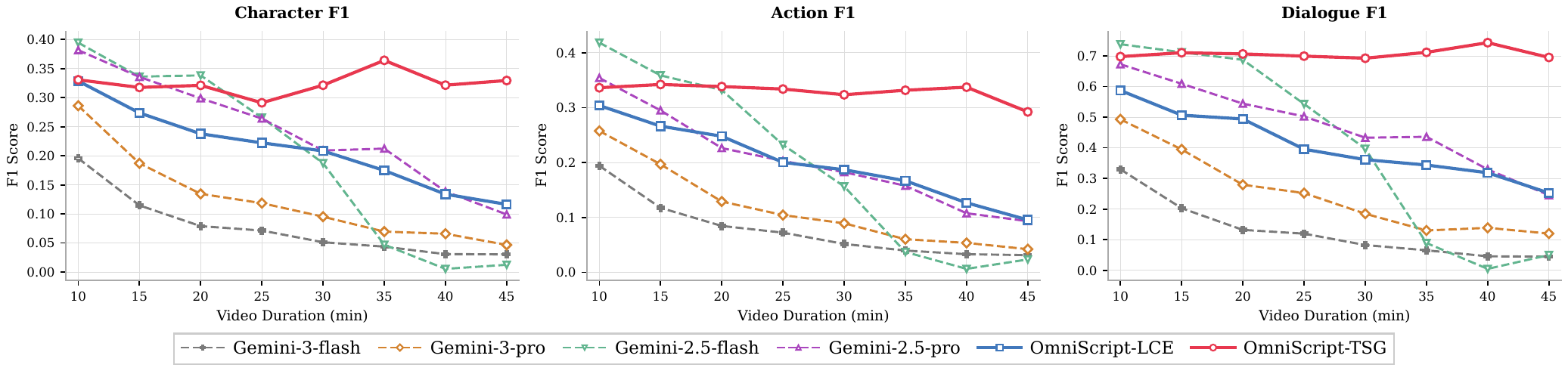}
        \caption{Event-level F1 scores of multiple field across long-video durations.}
    \end{subfigure}

    \begin{subfigure}[t]{\textwidth}
        \centering
        \includegraphics[width=\linewidth]{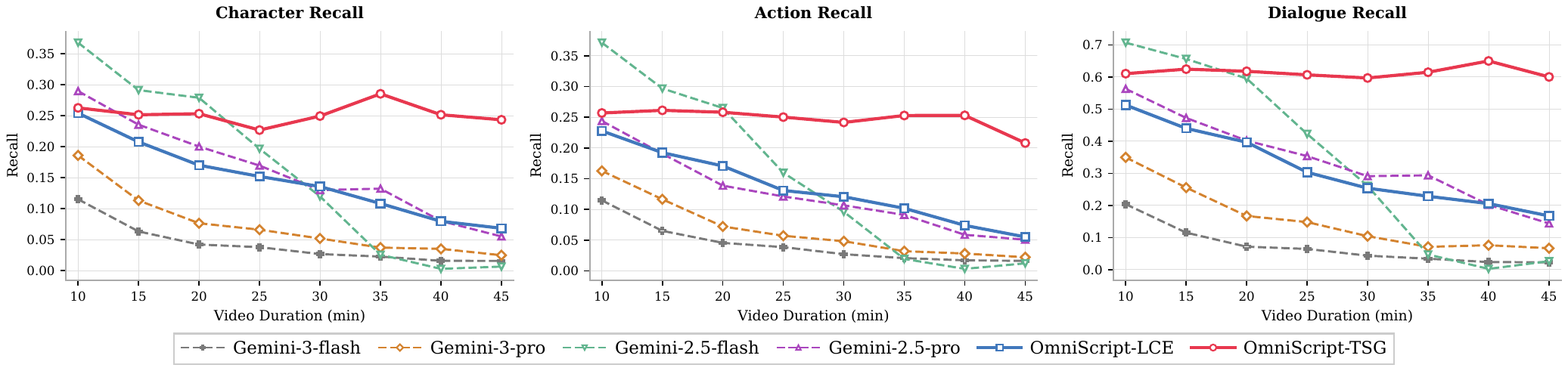}
        \caption{Event-level Recall scores of multiple field across long-video durations.}
    \end{subfigure}
    \caption{Additional long-video evaluation results. (a) Event-level, scene-level, and overall F1 scores across durations. (b) Event-level fine-grained F1 scores on character, action, and dialogue. (c) Event-level fine-grained Recall scores on character, action, and dialogue.}
    \label{fig:longvideo_pr_summary}
\end{figure*}

\noindent\textbf{Long-Form Video Understanding: Breaking the Context Barrier}

To systematically dissect the capabilities and limitations of multimodal foundation models over extended temporal horizons, we evaluate our proposed strategies (OmniScript-LCE and OmniScript-TSG) against strong baselines on long videos ranging from 10 to 45 minutes. We report a comprehensive suite of metrics, capturing both fine-grained event attributes (e.g., character consistency, dialogue, action) and scene-level structural alignment (e.g., temporal IoU, mood, environment).

\noindent\textbf{Multi-Dimensional Performance and the Context Cliff.}  As depicted in the radar charts (Figure~\ref{fig:longvideo_radar}), the multi-dimensional performance landscape undergoes a dramatic transformation as video duration scales. At shorter durations (10 and 20 minutes), the performance distributions of most models are relatively intertwined, forming expansive polygons. In this regime, OmniScript-LCE, benefiting from its end-to-end global context processing, achieves highly competitive multi-dimensional coverage alongside the strongest baseline, \texttt{Gemini-2.5-pro}. However, a critical context cliff emerges at the 30-minute mark. Models relying on standard global context assimilation, including our LCE variant and the Gemini baselines, experience a severe volumetric collapse across almost all 14 evaluated axes. This phenomenon exposes the devastating impact of accumulated long-range dependency errors and context dilution when processing continuous, hour-scale multimodal streams.

\noindent\textbf{The Inherent Bottleneck and Length-Induced Degeneration.}
This multi-dimensional collapse is further corroborated by the continuous trend lines in Figure~\ref{fig:longvideo_pr_summary}. There exists a pronounced negative correlation between video duration and generation quality for conventional architectures. Examining the Overall Avg F1 and Scene Avg F1 (Fig.~\ref{fig:longvideo_pr_summary}a), state-of-the-art baselines exhibit a severe, almost linear degradation trend. 
An intriguing anomaly emerges when analyzing the behavior of the \texttt{Gemini-2.5-flash} model. For video durations under 25 minutes, \texttt{Gemini-2.5-flash} achieves exceptionally high Event Avg F1 scores, surprisingly outperforming its more advanced counterparts, namely \texttt{Gemini-3-pro} and \texttt{Gemini-2.5-pro}. To unravel the underlying cause of this counter-intuitive superiority, we further visualize the Recall scores in Figure~\ref{fig:longvideo_pr_summary}(c). The data explicitly reveals that the high F1 performance of \texttt{Gemini-2.5-flash} on shorter videos is primarily driven by its markedly superior recall. This indicates that for video clips shorter than 25 minutes, the model tends to generate significantly more comprehensive and exhaustive outputs, successfully capturing a broader range of events compared to the more conservative generation patterns of other models.
However, a critical inflection point occurs as the video duration surpasses the 25-minute threshold, where the performance of \texttt{Gemini-2.5-flash} experiences a severe and precipitous decline. Through qualitative observation of the generated textual results at these extreme lengths, we identified that the model frequently degenerates into repetitive generation loops and produces unstructured outputs. This phenomenon clearly demonstrates that while the model exhibits comprehensive extraction capabilities on moderately long inputs, its sustained narrative comprehension and formatting stability over extended horizons are ultimately constrained by its inherent model capacity.

\noindent\textbf{Unprecedented Length-Invariance of the OmniScript-TSG Strategy.}
In stark contrast to the universal decay and volumetric collapse observed in all other models, our proposed OmniScript-TSG strategy demonstrates strong length-invariant robustness. As illustrated in Figure~\ref{fig:longvideo_radar}c and \ref{fig:longvideo_radar}d, while competitors shrink into small inner cores at 30 and 40 minutes, OmniScript-TSG maintains a massive, nearly unchanged polygonal area, dominating every single metric axis. This visual dominance translates perfectly to the flat trend lines in Figure~\ref{fig:longvideo_pr_summary}. Across almost all dimensions (i.e., Event, Scene, and Overall F1), the performance curve of OmniScript-TSG remains remarkably horizontal, fundamentally breaking the negative correlation between video length and generation quality. By enforcing segment-level conditioning that provides strict, localized constraints, TSG effectively insulates the generation process from global noise, permanently anchors narrative consistency, and bypasses the catastrophic forgetting typically induced by extreme sequence lengths.

\noindent\textbf{Moderate Degradation and the Role of OmniScript-LCE.}
While OmniScript-TSG serves as a robust solution for extreme lengths, our OmniScript-LCE strategy remains a highly effective alternative that balances extensive context processing with detailed semantic localization for moderate durations. Although LCE naturally suffers from the aforementioned multi-dimensional shrinkage at extreme lengths (Figure~\ref{fig:longvideo_radar}c/d), its performance decline is more moderate, exhibiting an aggregate metric decay rate (Figure~\ref{fig:longvideo_pr_summary}) that is noticeably slower than those of the baselines. Notably, on the video subsets long than 30 minutes, despite the volumetric loss, OmniScript-LCE still consistently matches or surpasses the rapidly degrading baselines across fine-grained metrics such as Character F1 and Action F1 (Fig.~\ref{fig:longvideo_pr_summary}b). This demonstrates that even under a global-context paradigm, the OmniScript framework offers better resilience against context dilution compared to other models.

\subsection{Ablation Studies}

\begin{table}[tbp]
  \centering
  \caption{Ablation study for training strategy.}
  \vspace{-0.5em}
  \label{tab:rl}
  \begin{tabular}{lccccccccc}
    \toprule
    \textbf{Model} & \textbf{CoT} & \textbf{Reward} & \textbf{Char.} & \textbf{Dia.} & \textbf{Act.} & \textbf{Exp.} & \textbf{Aud.} & \textbf{Overall} & \textbf{tIoU@0.1}  \\
    \midrule
    SFT & $\times$ & -- & 35.6 & 68.2 & 30.5 & 31.2 & 11.12 & 35.3 & 66.6 \\
    SFT & $\checkmark$ & -- & 37.8 & 71.0 & 33.5 & 31.2 & 11.5 & 37.0 & 68.9 \\
    SFT+RL  & $\times$ & Segmented & 37.1 & 70.9 & 32.8 & 32.5 & 11.7 & 37.0 & 69.0 \\
    SFT+RL  & $\checkmark$ & Global & 39.2 & 69.0 & 32.4 & 31.8 & 12.3 & 37.0 & 68.7 \\
    SFT+RL  & $\checkmark$ & Segmented & 39.2 &	72.2 &	33.7 &	31.9	& 11.6	& 37.7	& 69.3\\
    \bottomrule
  \end{tabular}
\end{table}

\noindent \textbf{Effectiveness of training strategy.} Table~\ref{tab:rl} ablates our Chain-of-Thought decoding and Reinforcement Learning (RL) stages. 
\textbf{Reasoning Trace (CoT):} Introducing CoT to the SFT baseline boosts the Overall score (35.3\% $\rightarrow$ 37.0\%) and Dialogue F1 score (68.2\% $\rightarrow$ 71.0\%). By constructing a latent cognitive scaffold prior to event decoding, CoT successfully enforces global-local consistency and significantly reduces long-context ambiguity.
\textbf{RL Alignment:} Transitioning to the RL stage further increases the Overall score to 37.7\%. By optimizing sequence-level metrics, RL effectively mitigates the exposure bias inherent in SFT auto-regressive decoding. 
\textbf{Segmented Reward:} Crucially, our proposed Segmented Reward outperforms the standard Global reward by rigorously penalizing fine-grained precision and recall errors at the event level.

\begin{table}[tbp]
  \centering
  \caption{Comparison of event-level performance with and without masking video subtitles. SV denotes Subtitle Visible ($\checkmark$: visible, $\times$: masked).}
  \label{tab:mask_subtitle_ablation}
  \begin{tabular}{lcccccccc}
    \toprule
    \textbf{Model} & \textbf{SV} & \textbf{Char.} & \textbf{Dialog} & \textbf{Action} & \textbf{Exp.} & \textbf{Audio} & \textbf{Overall} & \textbf{tIoU@0.1} \\
    \midrule
    Qwen3VL-235B-A22B & $\checkmark$ & 38.1 &	58.6 &	33.0 &	29.1 &	6.0 &	33.0	& 62.0 \\
    Qwen3VL-235B-A22B & $\times$ & 26.2 &	7.7 &	29.8 &	23.5 &	6.0 &	18.6 &	45.1 \\
    \midrule
    Gemini-3-pro & $\checkmark$ & 39.8	& 68.8	& 37.4	& 35.4	& 13.3	& 38.9	& 64.4 \\
    Gemini-3-pro & $\times$ & 40.4	& 60.9	& 34.7	& 33.6	& 13.2	& 36.6	& 60.3 \\
    \midrule
    \textbf{Ours-8B} & $\checkmark$ & 39.2 & 72.2 & 33.7 & 31.9 & 11.6 & 37.7 & 69.3 \\
    \textbf{Ours-8B} & $\times$ & 34.1 & 63.8 & 31.8 & 30.6 & 11.7 & 34.4 & 67.0 \\
    \bottomrule
  \end{tabular}
\end{table}

\noindent \textbf{Effectiveness of video subtitle.}
In Tables~\ref{tab:sota_event_level} and~\ref{tab:sota_scene_level}, several non-omni models (without audio input) still obtain relatively strong performance on dialogue F1 score, suggesting a potential shortcut from visible subtitles. To diagnose this effect, we mask subtitles at inference time and report the comparison in Table~\ref{tab:mask_subtitle_ablation}.
After subtitle masking, omni models such as Gemini-3-pro show a moderate degradation on dialogue F1 score (68.8 $\rightarrow$ 60.9), indicating that dialogue recognition is not purely text-copying and still relies on multimodal cues. This also suggests that omni models are more robust in subtitle-free settings. In contrast, Qwen3VL-235B-A22B exhibits a severe collapse (58.6 $\rightarrow$ 7.7), which suggests substantial dependence on visual information rather than robust audio-visual dialogue understanding.

\begin{table}[tbp]
  \centering
  \caption{Comparison of event-level performance with and without audio as input.}
  \label{tab:audio_ablation}
  \begin{tabular}{lcccccccc}
    \toprule
    \textbf{Model} & \textbf{Omni} & \textbf{Char.} & \textbf{Dia.} & \textbf{Act.} & \textbf{Exp.} & \textbf{Aud.} & \textbf{Overall} & \textbf{tIoU@0.1} \\
    \midrule
    Qwen3-VL-8B-SFT & $\times$ & 34.0 & 52.0 & 30.3 & 28.5 & 10.5 & 31.1 & 68.2 \\
    \textbf{Ours-8B-SFT} & $\checkmark$ & 35.6 & 68.2 & 30.5 & 31.2 & 11.1 & 35.3 & 66.6 \\
    \bottomrule
  \end{tabular}
\end{table}

\noindent\textbf{Effectiveness of Audio-Injection Pretraining.} 
To rigorously validate the necessity of the acoustic modality and our pretraining stage, we conduct an ablation study comparing our full model against a vision-only baseline. Specifically, we train a \texttt{Qwen3-VL-8B-SFT} model directly on our constructed dataset using only visual frames, completely depriving it of audio inputs. 
As illustrated in Table~\ref{tab:audio_ablation}, while the vision-only baseline manages to capture basic visual actions (scoring 30.3\% in Act.), it suffers a significant performance bottleneck in audio-dependent and implicit semantic fields. Most notably, the Dialogue recognition accuracy of the vision-only model is constrained to 52.0\%. In stark contrast, our full model, which leverages the audio-injection pretraining phase to align acoustic features with the LLM backbone, achieves a remarkable 68.2\% in Dialogue accuracy, yielding a massive absolute improvement of +16.2\%. 
This substantial performance gap underscores the fundamental limitation of vision-centric models in cinematic scripting and validates the effectiveness of our pretraining stage.

\subsection{Qualitative Analysis}

\begin{figure}[h]
    \centering
    \includegraphics[width=\linewidth]{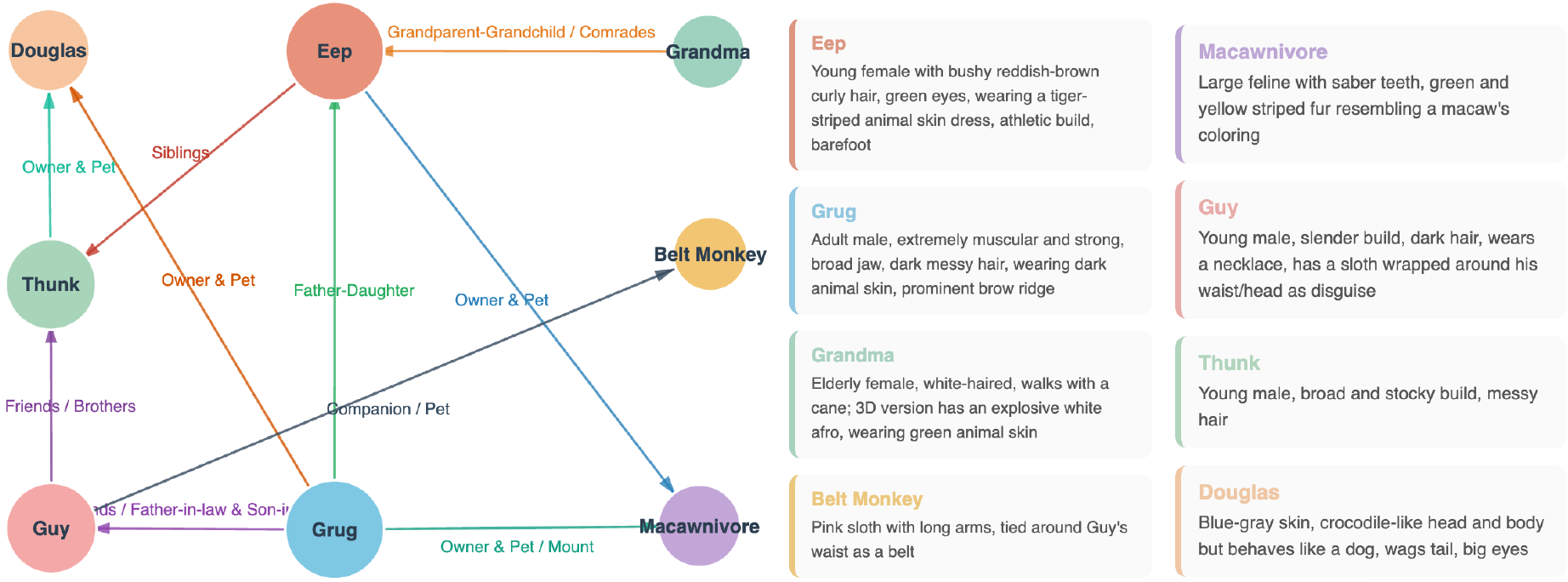}
    \caption{
        \textbf{Qualitative Visualization of the Character Profile Manager output (\textit{The Croods}).} 
        The generated profile consists of two core components. \textbf{Left:} A comprehensive character relationship graph detailing the social topology (e.g., siblings, parent-child, and companions). \textbf{Right:} Fine-grained appearance descriptions for each extracted identity. This structured representation serves as a persistent semantic memory, ensuring identity consistency and logically sound interactions during long-context script generation.
    }
    \label{fig:character_profile}
\end{figure}

\noindent \textbf{Character Profiles.} To intuitively demonstrate the efficacy of our proposed Character Profile Manager, we provide a qualitative visualization of the extracted character relationships and appearance attributes. Figure~\ref{fig:character_profile} illustrates a parsed profile generated from the animated film \textit{The Croods}. 

As detailed in the main text, the Character Profile Manager operates during the training data construction phase to establish a structured, persistent memory of the cast. The visualization highlights two critical dimensions of our extracted profiles that guide the script generation process:

\begin{itemize}
    \item \textbf{Relational Topology:} The character graph (left) captures the complex social dynamics within the video, accurately mapping familial ties (e.g., Father-Daughter between Grug and Eep), cross-generational relationships (e.g., Grandparent-Grandchild), and companion interactions (e.g., Owner \& Pet). This explicit topological structure equips the script generation model with strict relational constraints, ensuring that character interactions and dialogue remain logically consistent and contextually appropriate throughout the long-form narrative.
    
    \item \textbf{Fine-Grained Appearance Grounding:} The detailed appearance descriptions (right) showcase the granular visual understanding achieved by the manager. By explicitly capturing multidimensional attributes such as age, physique, clothing, and distinctive features (e.g., ``bushy reddish-brown curly hair'' for Eep, or the ``saber teeth'' of the Macawnivore), these descriptions serve as robust textual anchors. This effectively mitigates identity hallucination and ensures visual-semantic consistency when generating scenes involving open-vocabulary characters.
\end{itemize}

By seamlessly integrating these rich relational and visual priors, our approach significantly enhances the model's ability to maintain character faithfulness across thousands of generated words.

\begin{figure}
    \centering
    \includegraphics[width=0.95\textwidth]{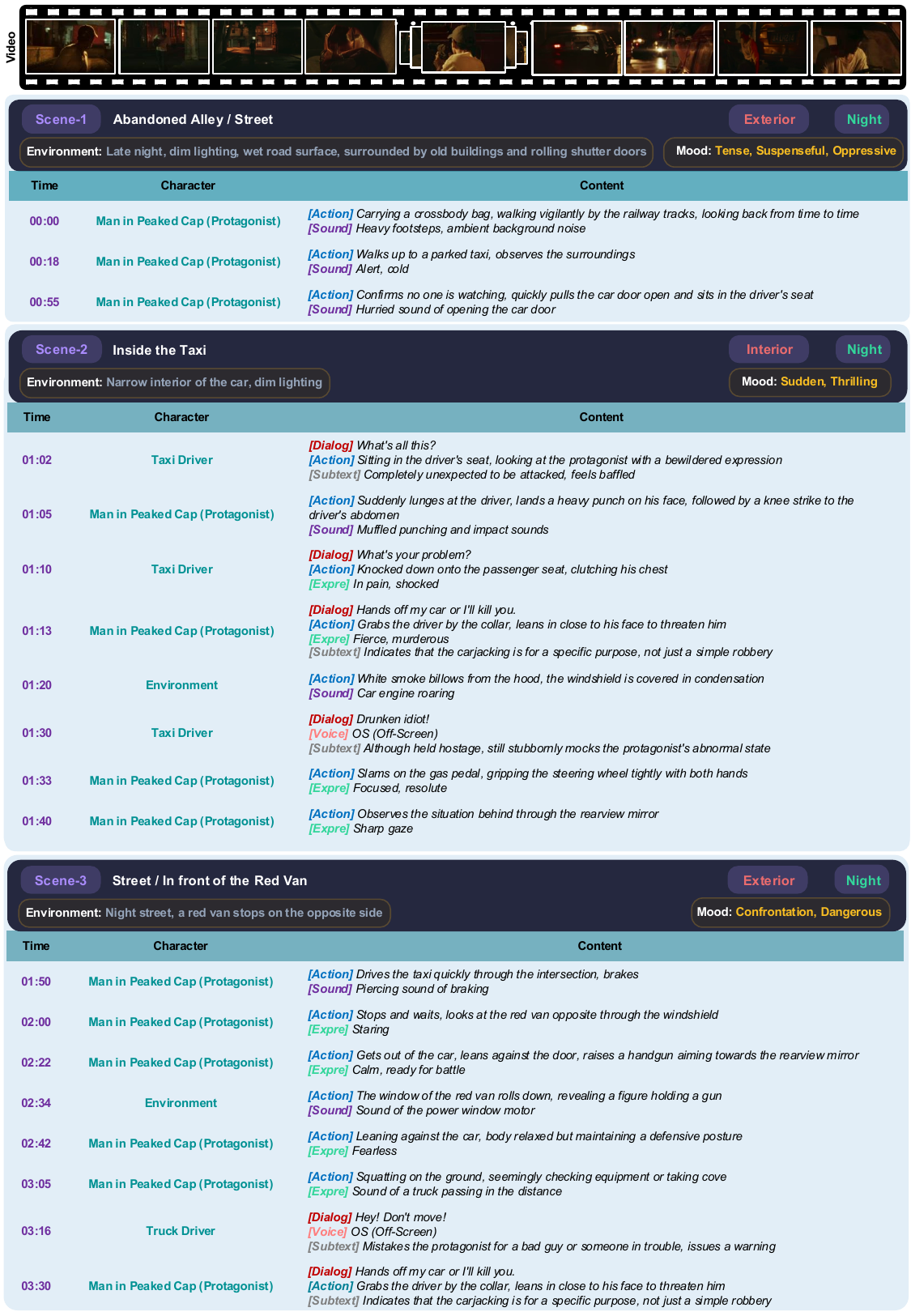}
    \vspace{-2mm}  
    \caption{\textbf{Qualitative visualization of OmniScript outputs.} OmniScript organizes each video into scene-level metadata and timestamped events with aligned characters, actions, dialogue, expressions, and audio cues, producing an interpretable and temporally coherent script representation.}
  \label{fig:example_vis} 
\end{figure}

\noindent \textbf{Script Generation.}  Fig.~\ref{fig:example_vis} illustrates the visualization of our structured video script transcription. 
The video is first segmented into scenes with global metadata such as environment and mood. 
Within each scene, the timeline is decomposed into timestamped events, where the system identifies the active characters and generates structured descriptions including actions, dialogue, sounds, expressions, and narrative cues. 
This hierarchical representation converts raw video into an interpretable story-centric structure, facilitating downstream tasks such as video understanding and retrieval.
\section{Conclusion}
\label{sec:conclusion}

In this paper, we present \textit{OmniScript}, an audio-visual language model for script-oriented understanding of long-form narrative videos. We target two complementary goals: accurate multi-field semantic parsing (event and scene attributes) and reliable temporal localization. We build the first comprehensive video script benchmarks with high-quality manual annotations, featuring long, complex cinematic videos. 
Extensive experiments on our benchmark show that OmniScript achieves a strong balance between quality and efficiency. With only 8B parameters, it consistently outperforms much larger open-source models on event-level understanding and temporal grounding, while remaining competitive on scene-level metrics. The subtitle-masking analysis further highlights robustness differences between model families and confirms the value of genuine audio-visual reasoning beyond text shortcuts.
Our findings suggest that robust script understanding still depends on better fine-grained multimodal perception, especially for subtle attributes and boundary-sensitive localization. We hope this work provides a useful baseline, benchmark, and set of training insights for future research on omni-modal narrative video understanding.

\clearpage
\bibliographystyle{assets/plainnat}
\bibliography{paper}

\newpage
\beginappendix
\section{Prompt Details}


\begin{promptbox}
You are annotating the first segment of a TV series/video. Please watch the video carefully and complete the following tasks:
\\[1em]
[Task Instructions]

1. **Scene Description**: Describe the main scenes in the video, including the time, location, and environmental atmosphere.

2. **Character Identification**: Identify all characters appearing in the video and record their physical appearances and vocal characteristics.

3. **Plot Summary**: Summarize the main plot development within this segment.

4. **Character Relationships**: Analyze the relationships between the characters.
\\[1em]
[Output Format]

Please output in JSON format, strictly adhering to the following structure:

```json \\
\{ \\
\hspace*{1em} "segment\_info": \{ \\
\hspace*{2em}        "segment\_index": 1, \\
\hspace*{2em}        "duration": "Video duration (e.g., 05:30)", \\
\hspace*{2em}        "main\_location": "Main scene location", \\
\hspace*{2em}        "time\_period": "Time of day (e.g., Daytime/Night/Morning)" \\
\hspace*{1em}    \}, \\
\hspace*{1em}    "characters": [ \\
\hspace*{2em}        \{ \\
\hspace*{3em}            "id": "char\_001", \\
\hspace*{3em}            "name": "Character name (if mentioned in the video) or 'Unnamed'", \\
\hspace*{3em}            "aliases": ["List of aliases"], \\
\hspace*{3em}            "appearance": "Detailed physical description: age, gender, height, body type, hairstyle, facial features, clothing, etc.", \\
\hspace*{3em}            "voice": "Vocal characteristics: tone, speaking rate, accent, etc.", \\
\hspace*{3em}            "personality": "Personality traits inferred from behavior and demeanor", \\
\hspace*{3em}            "first\_appear": "Timestamp of first appearance", \\
\hspace*{3em}            "role\_type": "Role type: Lead / Supporting / Background (Extra)", \\
\hspace*{3em}           "appearances\_in\_segment": [ \\
\hspace*{4em}                \{ \\
\hspace*{5em}                    "timestamp": "Appearance timestamp", \\
\hspace*{5em}                    "action": "Action description" \\
\hspace*{4em}                \} \\
\hspace*{3em}            ] \\
\hspace*{2em}        \} \\
\hspace*{1em}    ], \\
\hspace*{1em}    "plot\_summary": \{ \\
\hspace*{2em}        "main\_events": [ \\
\hspace*{3em}            \{ \\
\hspace*{4em}                "timestamp": "Event timestamp", \\
\hspace*{4em}                "description": "Event description" \\
\hspace*{3em}            \} \\
\hspace*{2em}        ], \\
\hspace*{2em}        "narrative": "Complete narrative of the plot (200-500 words)" \\
\hspace*{1em}    \}, \\
\hspace*{1em}    "relationships": [ \\
\hspace*{2em}        \{ \\
\hspace*{3em}            "char1\_id": "ID of character 1", \\
\hspace*{3em}            "char2\_id": "ID of character 2", \\
\hspace*{3em}            "relationship": "Relationship description (e.g., Spouses, Colleagues, Friends, Hostile, etc.)", \\
\hspace*{3em}            "evidence": "Basis for judgment / Evidence" \\
\hspace*{2em}        \} \\
\hspace*{1em}    ], \\
\hspace*{1em}    "scene\_transitions": [ \\
\hspace*{2em}        \{ \\
\hspace*{3em}            "timestamp": "Transition timestamp", \\
\hspace*{3em}            "from\_scene": "Original scene", \\
\hspace*{3em}            "to\_scene": "New scene", \\
\hspace*{3em}            "transition\_type": "Transition method (e.g., Cut, Dissolve, Fade in/out)" \\
\hspace*{2em}        \} \\
\hspace*{1em}    ], \\
\hspace*{1em}    "audio\_elements": \{ \\
\hspace*{2em}        "background\_music": "Background music description", \\
\hspace*{2em}        "sound\_effects": ["List of important sound effects"], \\
\hspace*{2em}        "ambient\_sound": "Ambient sound description" \\
\hspace*{1em}    \} \\
\} \\
'''
\\[1em]
[Important Notes]

- Character Naming Rule: If a character's name is explicitly mentioned in the video, use their real name; otherwise, use "Unnamed".

- Character ID Rule: Use char\_XXX for characters with known names, and unknown\_XXX for characters with unknown names.

- Appearance Description: Must be detailed and specific to facilitate the re-identification of the same character in subsequent video segments.

- Relationship Judgment: Must be supported by evidence from the video; avoid unfounded speculation.

Please begin analyzing the video content.
\end{promptbox}

\vspace{-1em}
\captionof{figure}{\textbf{Prompt used to process the first video clip in the character-centric plot reasoning stage.} This prompt instructs the LLM to identify characters, summarize the plot, and output the results in a strict JSON format.}
\label{fig:prompt_character_reasoning_1}

\vspace{1em}

\begin{promptbox}
You are annotating segment \{segment\_index\} of a TV series/video.

\{character\_context\}

[Plot Summary of Previous Segment]
\{prev\_segment\_summary\}
\\[1em]
[Task Instructions]

1. **Character Identification and Matching**:
   - Identify all characters appearing in the video.
   - Match them with known character profiles (based on appearance, voice, name, etc.).
   - If it is a known character, use their existing ID.
   - If it is a new character, create a new ID (Format: unknown\_XXX or char\_XXX).
   - If new information about a known character is discovered (e.g., the name of a previously unnamed character is mentioned), annotate the update.

2. **Scene Description**: Describe the main scenes in the video.

3. **Plot Summary**: Summarize the main plot development within this segment, paying attention to continuity with previous segments.

4. **Character Relationship Updates**: Update or supplement character relationship information.
\\[1em]
[Output Format]

Please output in JSON format, strictly adhering to the following structure:

```json \\
\{ \\
\hspace*{1em}"segment\_info": \{ \\
\hspace*{2em}"segment\_index": \{segment\_index\}, \\
\hspace*{2em}"duration": "Video duration", \\
\hspace*{2em}"main\_location": "Main scene location", \\
\hspace*{2em}"time\_period": "Time period description", \\
\hspace*{2em}"continuity\_note": "Continuity note regarding the previous segment" \\
\hspace*{1em}\}, \\
\hspace*{1em}"characters": [ \\
\hspace*{2em}\{ \\
\hspace*{3em}"id": "Character ID (use existing ID for known, generate new ID for new)", \\
\hspace*{3em}"name": "Character name or 'Unnamed'", \\
\hspace*{3em}"matched\_from": "Matched known character ID (if it is a known character)", \\
\hspace*{3em}"match\_confidence": "Match confidence: high/medium/low", \\
\hspace*{3em}"match\_reason": "Basis for matching", \\
\hspace*{3em}"new\_info": \{ \\
\hspace*{4em}"appearance": "Newly discovered physical traits (optional)", \\
\hspace*{4em}"voice": "Newly discovered vocal characteristics (optional)", \\
\hspace*{4em}"personality": "Newly discovered personality traits (optional)", \\
\hspace*{4em}"aliases": ["Newly discovered aliases"] \\
\hspace*{3em}\}, \\
\hspace*{3em}"appearances\_in\_segment": [ \\
\hspace*{4em}\{ \\
\hspace*{5em}"timestamp": "Appearance timestamp", \\
\hspace*{5em}"action": "Action description" \\
\hspace*{4em}\} \\
\hspace*{3em}] \\
\hspace*{2em}\} \\
\hspace*{1em}], \\
\hspace*{1em}"name\_updates": [ \\
\hspace*{2em}\{ \\
\hspace*{3em}"unknown\_id": "ID of previously unnamed character", \\
\hspace*{3em}"name": "Real name discovered in this segment", \\
\hspace*{3em}"evidence": "Evidence for the name (e.g., mentioned in dialogue)" \\
\hspace*{2em}\} \\
\hspace*{1em}], \\
\hspace*{1em}"plot\_summary": \{ \\
\hspace*{2em}"main\_events": [ \\
\hspace*{3em}\{ \\
\hspace*{4em}"timestamp": "Event timestamp", \\
\hspace*{4em}"description": "Event description", \\
\hspace*{4em}"involved\_characters": ["List of involved character IDs"] \\
\hspace*{3em}\} \\
\hspace*{2em}], \\
\hspace*{2em}"narrative": "Complete narrative of the plot", \\
\hspace*{2em}"connection\_to\_previous": "Connection to the previous plot" \\
\hspace*{1em}\}, \\
\hspace*{1em}"relationships": [ \\
\hspace*{2em}\{ \\
\hspace*{3em}"char1\_id": "ID of character 1", \\
\hspace*{3em}"char2\_id": "ID of character 2", \\
\hspace*{3em}"relationship": "Relationship description", \\
\hspace*{3em}"is\_new": true, \\
\hspace*{3em}"evidence": "Basis for judgment" \\
\hspace*{2em}\} \\
\hspace*{1em}], \\
\hspace*{1em}"new\_characters\_summary": [ \\
\hspace*{2em}\{ \\
\hspace*{3em}"id": "New character ID", \\
\hspace*{3em}"name": "Name or unnamed", \\
\hspace*{3em}"appearance": "Appearance description", \\
\hspace*{3em}"voice": "Vocal characteristics", \\
\hspace*{3em}"role\_in\_segment": "Role in this segment" \\
\hspace*{2em}\} \\
\hspace*{1em}] \\
\} \\
'''
\\[1em]
[Character Matching Guide]

- Prioritize matching by name (if mentioned in the video).
- Secondary matching by physical appearance (clothing, hairstyle, body type, etc.).
- Tertiary matching by vocal characteristics.
- Note that the same character may change clothes; comprehensive judgment is required.
\vspace{1em}
[Important Notes]

- Maintaining character ID consistency is the most important task.
- For uncertain matches, label the match\_confidence as low and explain the reason.
- When discovering new characters, provide as detailed a description as possible.
- Pay attention to plot continuity and remain consistent with previous segments.

Please begin analyzing the video content.
\end{promptbox}

\vspace{-1em}
\captionof{figure}{\textbf{Prompt used to process the subsequent video clips in the character-centric plot reasoning stage.} Building upon the initial segment, this prompt provides the LLM with previous character and plot context. It instructs the model to maintain temporal consistency by matching characters across segments, updating character profiles (e.g., discovering real names), and summarizing the continuing narrative in a structured JSON format.}
\label{fig:prompt_character_reasoning_2}

\begin{promptbox}

You are a senior screenwriter expert and a multimodal data annotator, as well as a computer vision expert proficient in cinematic language. You excel at reverse-engineering scripts by analyzing video visuals and audio (dialogue, intonation, sound effects, BGM). You can keenly capture the audience's psychological rhythm and identify the "High Points" (Gratification Points) in the show.

\vspace{1em}
[Task Objective]

Please analyze the uploaded video and its corresponding synopsis, and output a JSON dataset containing "Script Details" and "High-Point Analysis."

\vspace{1em}
[Core Instructions]

1. Script Structuring
Extract events from the video in chronological order.
\hspace*{1em}- **Field Independence**: `dialogue` and `action` must be stored separately.

\hspace*{1em}- **Non-empty Constraint**: For each Event object, **at least one** of the following fields must be present: "dialogue", "action", "expression", "audio\_cue". If a field does not exist at a given moment, omit it from the JSON directly (do not output "null" or empty strings).

\hspace*{1em}- **Concurrency**: If a character speaks while performing an action or showing a specific facial expression, include these fields within the same Event object.

\hspace*{1em}- **Semantic Depth**:

\hspace*{2em}- **Subtext**: If a line of dialogue contains irony, implication, or hidden intent, you must analyze and fill in the "subtext" field.

\hspace*{1em}- **Audio Details**: You must distinguish normal dialogue from off-screen voice, narration, etc.

\hspace*{1em}- **Accuracy**: Ensure that no character dialogue is missed. **This is extremely important, especially for long videos—pay special attention to ensuring that character dialogue is not omitted.**

2. Characters and Scenes
\hspace*{1em}- **Character Consistency**: Identify and maintain character IDs based on visual features (clothing, hairstyle) and voice characteristics.

\hspace*{1em}- **Scene Transitions**: A new scene should be created whenever the location, time, or **environment type (interior/exterior)** changes.

3. High-Point Mining
Identify segments in the video that produce strong emotional stimulation for the audience (e.g., successful revenge, truth revealed, ultimate romance, thrilling action, dark humor).

\hspace*{1em}- Your analysis must combine three dimensions: **Visual**, **Audio**, and **Text**.

\vspace{1em}
[Output Format (JSON Schema)]

Please strictly follow the JSON structure below. Do not output any content outside the Markdown code block.

```json \\
\{ \\
\hspace*{1em}"meta": \{ \\
\hspace*{2em}"title": "Video title or description", \\
\hspace*{2em}"duration": "Total video duration", \\
\hspace*{2em}"characters": ["Character A", "Character B"] \\
\hspace*{1em}\}, \\
\hspace*{1em}"script": [ \\
\hspace*{2em}\{ \\
\hspace*{3em}"scene\_id": 1, \\
\hspace*{3em}"location": "Scene location (e.g., cable car cabin)", \\
\hspace*{3em}"type": "Interior / Exterior", \\
\hspace*{3em}"environment": "Detailed description of the scene setting (e.g., an old cabin with a river view outside the window)", \\
\hspace*{3em}"time": "Day / Night / Early Morning / Dawn / Dusk / Noon, etc.", \\
\hspace*{3em}"mood": "Atmosphere of the scene (e.g., absurd, oppressive, tense)", \\
\hspace*{3em}"events": [ \\
\hspace*{4em}\{ \\
\hspace*{5em}"timestamp": "MM:SS", \\
\hspace*{5em}"character": "Character name (if it is ambient sound, use 'Environment')", \\
\\
\hspace*{5em}// --- Visual and Actions --- \\
\hspace*{5em}"action": "Specific physical action (e.g., throws a soda can out the window)", \\
\hspace*{5em}"expression": "Facial expression (e.g., intoxicated, vacant gaze)", \\
\\
\hspace*{5em}// --- Audio and Dialogue --- \\
\hspace*{5em}"dialogue": "Dialogue content (accurate transcription)", \\
\hspace*{5em}"voice\_type": "Normal (regular dialogue) / VO (voice-over – inner monologue) / OS (off-screen voice – character off screen) / NA (narration) [omit this field for normal dialogue]", \\
\hspace*{5em}"audio\_cue": "Non-verbal sound cues (e.g., a crisp slap sound)", \\
\\
\hspace*{5em}// --- Semantic Understanding --- \\
\hspace*{5em}"subtext": "Subtext (e.g., appears to be praise but is actually teasing) [optional]" \\
\hspace*{4em}\} \\
\hspace*{3em}] \\
\hspace*{2em}\} \\
\hspace*{1em}], \\
\hspace*{1em}"high\_points": [ \\
\hspace*{2em}\{ \\
\hspace*{3em}"id": 1, \\
\hspace*{3em}"type": "High-point type (emotional release / twist / dark humor / action)", \\
\hspace*{3em}"time\_range": ["Start time", "End time"], \\
\hspace*{3em}"description": "Brief summary of the key events in this segment", \\
\hspace*{3em}"reasoning": \{ \\
\hspace*{4em}"visual": "Visual stimulation (e.g., rapid cuts combined with punch impacts)", \\
\hspace*{4em}"audio": "Audio coordination (e.g., music abruptly stops)", \\
\hspace*{4em}"text": "Key dialogue", \\
\hspace*{4em}"psychology": "Source of audience gratification (e.g., an arrogant character getting instantly humiliated)" \\
\hspace*{3em}\}, \\
\hspace*{3em}"score": 9.5 \\
\hspace*{2em}\} \\
\hspace*{1em}] \\
\} \\
'''

\vspace{1em}
To help you understand, here is an ideal example of an Event output:

Ideal JSON Event Output:

```json \\
\{ \\
\hspace*{1em}"timestamp": "00:45", \\
\hspace*{1em}"character": "Xie Xiaomeng", \\
\hspace*{1em}"action": "Leans against the window and gently shakes a drink", \\
\hspace*{1em}"expression": "Affectionate, self-absorbed", \\
\hspace*{1em}"dialogue": "Do you know what quality of yours attracts me? Melancholy!", \\
\hspace*{1em}"voice\_type": "VO (voice-over – inner monologue)", \\
\hspace*{1em}"subtext": "Attempting to appear profound in order to flirt with a woman, revealing his pretentious taste.", \\
\hspace*{1em}"audio\_cue": "Low-frequency rumble of the cable car in motion" \\
\} \\
'''

\vspace{1em}
[Plot Synopsis Corresponding to the Video]

<Synopsis of your video>

\end{promptbox}

\vspace{-1em}
\captionof{figure}{\textbf{Prompt used for video-to-script generation and high-point analysis with the corresponding synopsis.} This prompt guides the LLM to produce fine-grained, temporally ordered event annotations and explicit high-point reasoning in a structured JSON format.}
\label{fig:prompt_videoplot2script}
\vspace{-1em}

\vspace{1em}

\subsection{Character-centric Plot Reasoning}
In our character-centric plot reasoning framework, the Large Language Model (LLM) acts as the core reasoning engine. To ensure the LLM generates accurate, consistent, and easily parsable outputs across long videos, we meticulously designed two distinct prompts. These prompts instruct the LLM to output results in a highly structured JSON format, effectively transforming unstructured video content into a structured temporal database.

\vspace{1em}
\noindent\textbf{1. Initialization Prompt (First Segment)} \\
As shown in Fig.~\ref{fig:prompt_character_reasoning_1}, the initialization prompt is designed to tackle the ``cold start'' problem of a video. Since no prior knowledge exists, the LLM is instructed to perform a dense analysis of the scene, characters, and initial plot. 
A critical design choice in this prompt is the Character ID Rule. To handle situations where a character's name is not immediately revealed in the dialogue, we force the LLM to assign explicit identifiers: \texttt{char\_XXX} for characters with explicitly mentioned names, and \texttt{unknown\_XXX} for unnamed characters. Furthermore, the LLM is required to generate fine-grained multi-modal descriptions for each character (e.g., appearance, voice, inferred personality). These detailed descriptions serve as the foundational anchors for cross-segment character matching in subsequent steps.

\vspace{1em}
\noindent\textbf{2. Tracking and Reasoning Prompt (Subsequent Segments)} \\
The prompt shown in Fig.~\ref{fig:prompt_character_reasoning_2} is applied to all subsequent segments to maintain temporal consistency. Unlike the initialization prompt, this prompt is dynamically constructed by injecting the historical context (\texttt{\{character\_context\}} and \texttt{\{prev\_segment\_summary\}}). 

This prompt is specifically engineered to perform continuous character tracking and profile updating. We designed several specific JSON fields to guide the LLM's reasoning process:
\begin{itemize}
    \item \texttt{match\_confidence} and \texttt{match\_reason}: Instead of blindly assigning IDs, the LLM must explicitly state its confidence level and visual/acoustic reasoning for matching a detected character to the existing profile memory.
    \item \texttt{name\_updates}: This field elegantly solves the delayed-naming issue in TV series. If a character previously tracked as \texttt{unknown\_XXX} is finally addressed by name, the LLM captures this evidence to update the global profile.
    \item \texttt{continuity\_note}: This forces the LLM to explicitly reason about how the current events connect to the injected \texttt{\{prev\_segment\_summary\}}, minimizing hallucinations and ensuring a coherent plot narrative.
\end{itemize}
By explicitly constraining the output schema and providing specific matching guidelines (e.g., prioritizing names, then appearance, then voice), these prompts enable the LLM to reliably extract complex character dynamics without losing track of identities over long temporal windows.

\vspace{1em}
\noindent\textbf{3. Video-to-Script Prompt with Synopsis} \\
Given a video and its plot synopsis, this prompt instructs the LLM to generate temporally ordered script events and identify key high points in a unified JSON output. The synopsis serves as global narrative guidance to improve long-range coherence and reduce identity or event inconsistencies.


\subsection{Evaluation Prompts}

\begin{promptbox}
You are an expert video script evaluator specializing in \{field\_name\} matching.

Compare the Ground Truth (GT) with the Prediction (Pred) and provide a similarity score.

Ground Truth: "\{gt\_text\}"

Prediction: "\{pred\_text\}"

\{matching\_criteria\}

\{scoring\_guide\}

Task: Rate the semantic similarity between the GT and Pred descriptions.

Output strictly in JSON format:
\{

\hspace*{1em}    "similarity": <float between 0.0 and 1.0>,

\hspace*{1em}    "reason": "Brief explanation in Chinese"

\}

\end{promptbox}

\vspace{-1em}
\captionof{figure}{\textbf{Prompt template used to evaluate the field content.}}
\label{fig:prompt_template}
\vspace{1em}

\begin{promptbox}
Criteria for matching:

- Focus on whether the prediction captures the general scene/situation.

- Allow significant paraphrasing, different perspectives, or partial descriptions.

- As long as the prediction is not contradictory and describes a related action/situation, consider it correct.

- Different granularity levels are acceptable (e.g., detailed vs. summary).

- Example: GT="Zhang San walked quickly to the door, pushed it open and went out" vs Pred="Zhang San left" -> Correct (Generalized description)

- Example: GT="Zhang San slammed the table angrily" vs Pred="Zhang San is very angry" -> Correct (Emotion-related)

\vspace{1em}
Scoring Guidance:

- 1.0: Match - describes the same general situation/scene

- 0.7-0.9: Good match - captures the essence, may differ in detail/perspective

- 0.4-0.6: Acceptable - related situation, different granularity

- 0.1-0.3: Marginal - tangentially related

- 0.0: No match - contradictory description
\end{promptbox}

\vspace{-1em}
\captionof{figure}{\textbf{Matching criteria and scoring guidance used to evaluate the field content of Action.}}
\label{fig:prompt_action}
\vspace{1em}

\begin{promptbox}
Criteria for matching:

- Focus on whether the prediction captures the general audio atmosphere/category.

- Allow significant paraphrasing, different perspectives, or partial descriptions.

- As long as the prediction is not contradictory and describes related audio elements, consider it correct.

- Example: GT="Sound of breaking glass accompanied by a scream" vs Pred="Sound of breaking glass" -> Correct (Partial match)

- Example: GT="Tense background music" vs Pred="Soundtrack" -> Correct (Generalized description)

\vspace{1em}
Scoring Guidance:

- 1.0: Match - describes the same general audio atmosphere

- 0.7-0.9: Good match - captures the essence, may differ in detail

- 0.4-0.6: Acceptable - related audio, different granularity

- 0.1-0.3: Marginal - tangentially related

- 0.0: No match - contradictory description
\end{promptbox}

\vspace{-1em}
\captionof{figure}{\textbf{Matching criteria and scoring guidance used to evaluate the field content of Audio Cue.}}
\label{fig:prompt_audio}
\vspace{1em}

\begin{promptbox}
Criteria for matching:

- Focus on whether the prediction captures the general emotional valence (positive/negative/neutral).

- Allow significant paraphrasing, different granularity, or partial descriptions.

- As long as the prediction is not contradictory in emotional direction, consider it correct.

- Example: GT="Anger, disappointment, pain" vs Pred="Negative emotion" -> Correct (Generalized description)

- Example: GT="Joy, excitement" vs Pred="Happy" -> Correct (Positive emotion)

\vspace{1em}
Scoring Guidance:

- 1.0: Match - same emotional valence (positive/negative/neutral)

- 0.7-0.9: Good match - captures the general emotional tone

- 0.4-0.6: Acceptable - related emotional state, different granularity

- 0.1-0.3: Marginal - tangentially related emotions

- 0.0: No match - opposite emotional valence
\end{promptbox}

\vspace{-1em}
\captionof{figure}{\textbf{Matching criteria and scoring guidance used to evaluate the field content of Expression.}}
\label{fig:prompt_expression}
\vspace{1em}

\begin{promptbox}
Criteria for matching:

- The prediction should capture the main location type and general setting.

- Allow reasonable paraphrasing and minor detail differences.

- Example: GT="Zhang San's bedroom" vs Pred="Bedroom" -> Correct (Core location consistent)

- Example: GT="Hospital corridor" vs Pred="Inside the hospital" -> Correct (Same broader scene)

\vspace{1em}
Scoring Guidance:

- 1.0: Excellent match - same location, may have minor detail differences

- 0.7-0.9: Good match - core location matches, reasonable paraphrasing

- 0.4-0.6: Partial match - same general area/building

- 0.1-0.3: Weak match - loosely related location

- 0.0: No match - completely different location
\end{promptbox}

\vspace{-1em}
\captionof{figure}{\textbf{Matching criteria and scoring guidance used to evaluate the field content of Scene Location.}}
\label{fig:prompt_scene_location}
\vspace{1em}

\begin{promptbox}
Criteria for matching:

- Same as strict for this field, as type is categorical.

- Synonyms are allowed: "Interior" == "Int", "Exterior" == "Ext"

\vspace{1em}
Scoring Guidance:

- 1.0: Match - same scene type (synonyms allowed)

- 0.0: No match - different scene type
\end{promptbox}

\vspace{-1em}
\captionof{figure}{\textbf{Matching criteria and scoring guidance used to evaluate the field content of Scene Type.}}
\label{fig:prompt_scene_type}
\vspace{1em}

\begin{promptbox}
Criteria for matching:

- The prediction should capture the main environmental characteristics.

- Allow missing minor details if the core atmosphere is correct.

- Example: GT="Dim lighting, shabby furniture" vs Pred="Dim and shabby environment" -> Correct

\vspace{1em}
Scoring Guidance:

- 1.0: Excellent match - same environmental atmosphere

- 0.7-0.9: Good match - core environment matches

- 0.4-0.6: Partial match - related atmosphere

- 0.1-0.3: Weak match - loosely related

- 0.0: No match - contradictory environment
\end{promptbox}

\vspace{-1em}
\captionof{figure}{\textbf{Matching criteria and scoring guidance used to evaluate the field content of Scene Environment.}}
\label{fig:prompt_scene_env}
\vspace{1em}

\begin{promptbox}
Criteria for matching:

- Same as strict for this field, as time is categorical.

- Allow synonyms and reasonable time period mappings.

\vspace{1em}
Scoring Guidance:

- 1.0: Match - same time period (synonyms allowed)

- 0.0: No match - different time period
\end{promptbox}

\vspace{-1em}
\captionof{figure}{\textbf{Matching criteria and scoring guidance used to evaluate the field content of Scene Time.}}
\label{fig:prompt_scene_time}
\vspace{1em}

\begin{promptbox}
Criteria for matching:

- The prediction should capture the main mood and its general tone.

- Allow missing minor mood elements if the core atmosphere is correct.

- Example: GT="Tense, oppressive" vs Pred="Tense and uneasy" -> Correct (Core mood consistent)

Scoring Guidance:

- 1.0: Excellent match - same mood tone

- 0.7-0.9: Good match - core mood matches

- 0.4-0.6: Partial match - related emotional tone

- 0.1-0.3: Weak match - loosely related

- 0.0: No match - contradictory mood
\end{promptbox}

\vspace{-1em}
\captionof{figure}{\textbf{Matching criteria and scoring guidance used to evaluate the field content of Scene Mood.}}
\label{fig:prompt_scene_mood}
\vspace{1em}

Evaluating open-ended video script generation using traditional n-gram-based metrics (e.g., BLEU, ROUGE) is inherently flawed, as they heavily penalize valid paraphrasing and differing levels of descriptive granularity. To robustly evaluate the quality of the generated scripts, we employ LLM to evaluate the similarity. We design specific evaluation prompts for different script fields to assess the semantic equivalence between the Ground Truth (GT) and the Predictions (Pred). 

The core philosophy of our evaluation is to focus on semantic correctness and contextual alignment rather than strict lexical overlap. We categorize the evaluation criteria into three main types based on the nature of the fields:

\noindent\textbf{1. Semantic and Granularity Matching (Action, Audio Cue, Expression, Scene Mood)} \\
For highly subjective and dynamic fields, human annotators often describe the same event at varying levels of detail. Our prompts (Fig.~\ref{fig:prompt_action}, \ref{fig:prompt_audio}, \ref{fig:prompt_expression}, and \ref{fig:prompt_scene_mood}) explicitly instruct the LLM to tolerate significant paraphrasing, different perspectives, and partial matches. A prediction is deemed correct as long as it captures the general situation or core emotional valence without introducing contradictory information. For instance, in the \textit{Expression} field, predicting a generalized ``Negative emotion'' for a detailed GT of ``Anger, disappointment, pain'' is evaluated as a correct match.

\noindent\textbf{2. Spatial and Environmental Alignment (Scene Location, Scene Environment)} \\
For spatial and environmental descriptions, the evaluation focuses on the core setting. As shown in Fig.~\ref{fig:prompt_scene_location} and Fig.~\ref{fig:prompt_scene_env}, the LLM judge allows for reasonable spatial hierarchies (e.g., predicting the broader scene ``Inside the hospital'' for the GT ``Hospital corridor'') and forgives missing minor background details, provided the main environmental atmosphere is accurately captured.

\noindent\textbf{3. Categorical Synonym Matching (Scene Type, Scene Time)} \\
For fields with a more constrained vocabulary, the matching criteria are stricter but remain robust to synonyms and semantic mappings. As detailed in Fig.~\ref{fig:prompt_scene_type} and Fig.~\ref{fig:prompt_scene_time}, the LLM handles industry-standard abbreviations (e.g., matching ``Ext'' with ``Exterior'') and reasonable time-period mappings, ensuring that formatting differences do not result in false penalties.

By employing these customized, field-specific prompts, our evaluation protocol effectively handles the intrinsic variance of video-to-text generation, yielding metrics that correlate much closer with human judgment.

To implement the character mapping mechanism described in the main paper, we carefully engineered the alignment prompt shown in Fig.~\ref{fig:prompt_character_matching}. Rather than relying on simple text similarity, this prompt guides the LLM through a rigorous, five-step logical pipeline to resolve character ambiguities and output a structured JSON mapping:

\begin{itemize}
    \item \textbf{Categorization (Tasks 1 \& 2):} The LLM first classifies all detected entities into proper names versus identity names, and further distinguishes between singular and plural identities. This foundational step prevents basic mapping errors, such as incorrectly aligning a specific individual with a group entity (e.g., ``soldiers'').
    \item \textbf{Alias Resolution (Task 3):} The prompt explicitly instructs the LLM to handle many-to-one mappings. It identifies different title variants (e.g., ``Old Jin'' and ``Boss Jin'') and clusters them into the correct Ground Truth identity without forcing matches between distinct proper names.
    \item \textbf{Strict Conflict Detection (Tasks 4 \& 5):} Instead of merely calculating semantic similarity, the prompt acts as a logical filter. Task 4 detects direct identity incompatibilities based on gender or opposing roles (e.g., ``Police'' vs. ``Thief''). Crucially, Task 5 introduces \textit{cross-type conflict detection}, which prevents logical contradictions during fallback matching (e.g., forbidding the proper name ``Dr. Wang'' from being matched to the identity ``Security Guard'').
\end{itemize}

By enforcing this multi-step reasoning process and requiring a comprehensive JSON output, the prompt ensures that our evaluation script can deterministically parse the resolved aliases and safely apply the fallback matching strategies.

\vspace{2em}


\begin{promptbox}
You are an expert in character name classification and matching. Please complete all the following tasks:
\\[1em]
\#\# Input
Ground Truth Character List: [\{gt\_str\}] \\
Prediction Character List: [\{pred\_str\}]
\\[1em]
\#\# Task 1: Character Name Classification
Categorize the character names in both lists into two categories:

**Proper Names (proper\_names)**: Real human names, including:

- Chinese/Personal names (e.g., Zhang San, Li Ming, Jin Runfa)

- Title variants (e.g., Old Jin, Mr. Jin, Boss Jin are all variants of a person's name)

- Foreign names (e.g., John, Mary)

- Names with surname + occupation (e.g., Dr. Wang, Teacher Zhang, Officer Li) — although these contain occupational info, they are still classified as proper names.

**Identity Names/Pronouns (identity\_names)**: Non-name character identifiers, including:

- Occupational identities (e.g., soldier, nurse, doctor, police)

- Social roles (e.g., passerby, customer, boss)

- Descriptive pronouns (e.g., fatty, tall guy, lady, old man)

- Relational terms (e.g., mom, grandpa, unless explicitly used as a proper name)

- Numbered characters (e.g., Soldier A, Passerby B)
\\[1em]
\#\# Task 2: Identity Name Plurality Classification

For the identity names categorized in Task 1, further distinguish between **singular identity names** and **plural identity names**:

**Singular Identity Names (singular\_identities)**: Refers to a single character.

- Examples: soldier, nurse, doctor, police, passerby, lady, old man, fatty

**Plural Identity Names (plural\_identities)**: Refers to multiple characters (a group).

- With plural suffixes: soldiers, nurses, students, audiences

- With quantifiers: the crowd, a group of people, two soldiers

- Group vocabulary: crowd, masses, everyone, all soldiers

- With 'et al.' or 'and others' suffix: Zhang San and others
\\[1em]
\#\# Task 3: Proper Name Matching

For the classified proper names, determine which proper name in the Prediction refers to the same person as a proper name in the GT (could be aliases/title variants).

Matching rules:

- The same person might have different titles: Old Jin, Mr. Jin, Boss Jin, Jin Runfa $\rightarrow$ Same person

- Surname + occupation variants: Dr. Wang vs. Physician Wang $\rightarrow$ Likely the same person

- **Allow many-to-one**: Multiple Pred names can match the same GT name.

- Do not forcibly match different proper names (Zhang San vs. Li Si $\rightarrow$ Different people).
\\[1em]
\#\# Task 4: Identity Name Conflict Detection
For the classified identity names, detect which Prediction identity names are **incompatible** (cannot be the same character) with which GT identity names.

Incompatibility judgment rules:

1. **Gender conflict**: Lady vs. Gentleman $\rightarrow$ Incompatible

2. **Opposing identities**: Police vs. Thief, Customer vs. Waiter $\rightarrow$ Incompatible

3. **Inconsistent function/domain**: Security Guard vs. Nurse, Driver vs. Chef $\rightarrow$ Incompatible

Compatible situations (not considered conflicts):

- Same domain, different levels: Soldier vs. Officer

- Same domain, different positions: Nurse vs. Doctor

- Generic vs. Specific: Lady vs. Nurse

- Synonyms: Passerby vs. Pedestrian
\\[1em]
\#\# Task 5: Cross-type Conflict Detection (Important!)

Detect occupational conflicts between **proper names containing occupational info** and **identity names**.

Examples:

- "Dr. Wang" (Proper name, occupation=Doctor) vs. "Police" (Identity name) $\rightarrow$ Incompatible (Doctor $\neq$ Police)

- "Teacher Zhang" (Proper name, occupation=Teacher) vs. "Nurse" (Identity name) $\rightarrow$ Incompatible (Teacher $\neq$ Nurse)

- "Officer Li" (Proper name, occupation=Police) vs. "Security Guard" (Identity name) $\rightarrow$ Compatible (Both in security domain)

- "Dr. Wang" (Proper name, occupation=Doctor) vs. "Nurse" (Identity name) $\rightarrow$ Compatible (Both in medical domain)

This detection prevents fallback matching from incorrectly mapping occupational proper names to unrelated identity names.
\\[1em]

\#\# Output Format
Please strictly output in the following JSON format:

```json \\
\{ \\
\hspace*{1em}"gt\_classification": \{ \\
\hspace*{2em}"proper\_names": ["List of GT proper names"], \\
\hspace*{2em}"identity\_names": ["List of GT identity names"] \\
\hspace*{1em}\}, \\
\hspace*{1em}"pred\_classification": \{ \\
\hspace*{2em}"proper\_names": ["List of Pred proper names"], \\
\hspace*{2em}"identity\_names": ["List of Pred identity names"] \\
\hspace*{1em}\}, \\
\hspace*{1em}"gt\_identity\_plurality": \{ \\
\hspace*{2em}"singular": ["GT singular identity names"], \\
\hspace*{2em}"plural": ["GT plural identity names"] \\
\hspace*{1em}\}, \\
\hspace*{1em}"pred\_identity\_plurality": \{ \\
\hspace*{2em}"singular": ["Pred singular identity names"], \\
\hspace*{2em}"plural": ["Pred plural identity names"] \\
\hspace*{1em}\}, \\
\hspace*{1em}"proper\_name\_mappings": \{ \\
\hspace*{2em}"pred\_proper\_name\_1": "matched\_gt\_proper\_name\_or\_null", \\
\hspace*{2em}"pred\_proper\_name\_2": "matched\_gt\_proper\_name\_or\_null" \\
\hspace*{1em}\}, \\
\hspace*{1em}"identity\_conflicts": \{ \\
\hspace*{2em}"pred\_identity\_1": ["incompatible\_gt\_identity\_1", "incompatible\_gt\_identity\_2"], \\
\hspace*{2em}"pred\_identity\_2": ["incompatible\_gt\_identity\_3"] \\
\hspace*{1em}\}, \\
\hspace*{1em}"cross\_type\_conflicts": \{ \\
\hspace*{2em}"pred\_proper\_name\_with\_occupation": ["incompatible\_gt\_identity\_1", "incompatible\_gt\_identity\_2"], \\
\hspace*{2em}"gt\_proper\_name\_with\_occupation": ["incompatible\_pred\_identity\_1"] \\
\hspace*{1em}\} \\
\} \\
[1em]
Notes:

- All character names must appear in the classification results (proper\_names or identity\_names).

- All identity names must appear in the plurality classification (singular or plural).

- proper\_name\_mappings only contains Pred's proper names, with values being the matched GT name or null.

- identity\_conflicts only contains conflicts between Pred identity names and GT identity names.

- cross\_type\_conflicts contains conflicts between occupational proper names and opposing identity names (bidirectional detection):

- Pred occupational proper names vs. GT identity names

- GT occupational proper names vs. Pred identity names
\end{promptbox}

\vspace{-1em}
\captionof{figure}{\textbf{Prompt used for Character Mapping.} This prompt instructs the LLM to structurally classify names, resolve aliases, and explicitly detect semantic and cross-type domain conflicts to ensure accurate script evaluation.}
\label{fig:prompt_character_matching}
\section{Dataset Statistics}

\subsection{Trainingset Statistics}

\begin{figure}[h]
    \centering
    \includegraphics[width=\textwidth]{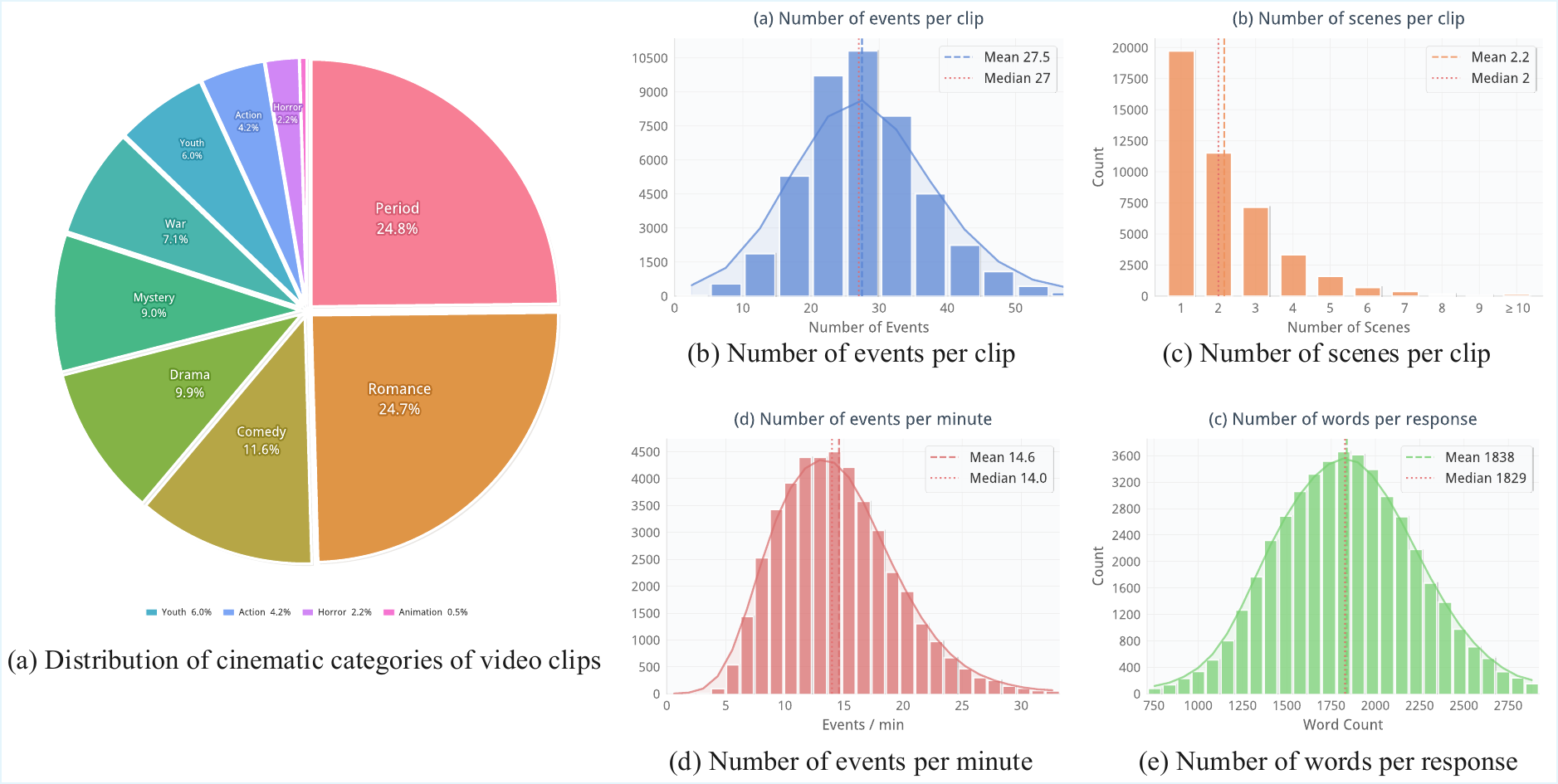}
    \caption{Statistics of training set for our OmniScript model.
    }
    \label{fig:train_statistics}
    
\end{figure}

Our training set is compiled from two complementary sources: a collection of long-form TV dramas with synopsis-grounded video clips, and a corpus of short vertical dramas popular on streaming platforms. 
Together they yield $45k$ clips spanning 781 unique titles across 10 cinematic genres (Fig.~\ref{fig:train_statistics}(a)), with period and romance dramas each accounting for roughly a quarter of the data, followed by comedy (11.6\%), drama (9.9\%), and mystery (9.0\%). 
This breadth ensures the model is exposed to diverse narrative structures, pacing styles, and visual conventions.

Each clip is paired with a structured multimodal script annotation produced by a vision-language model and subsequently filtered for quality. 
The annotation comprises a per-clip scene decomposition and a flat list of timestamped events, where each event records the character, action, dialogue, expression, and audio cue at that moment. 
On average, a clip contains 27.5 events across 2.2 scenes at a density of 14.6 events per minute, with responses averaging 1,838 words (Fig.~\ref{fig:train_statistics}(b–e)). 
The near-identical mean and median values across all four metrics confirm that the distribution is well-concentrated with few pathological outliers, reflecting consistent annotation quality across sources.

\subsection{Benchmark Statistics}

\begin{figure}[h]
    \centering
    \includegraphics[width=\textwidth]{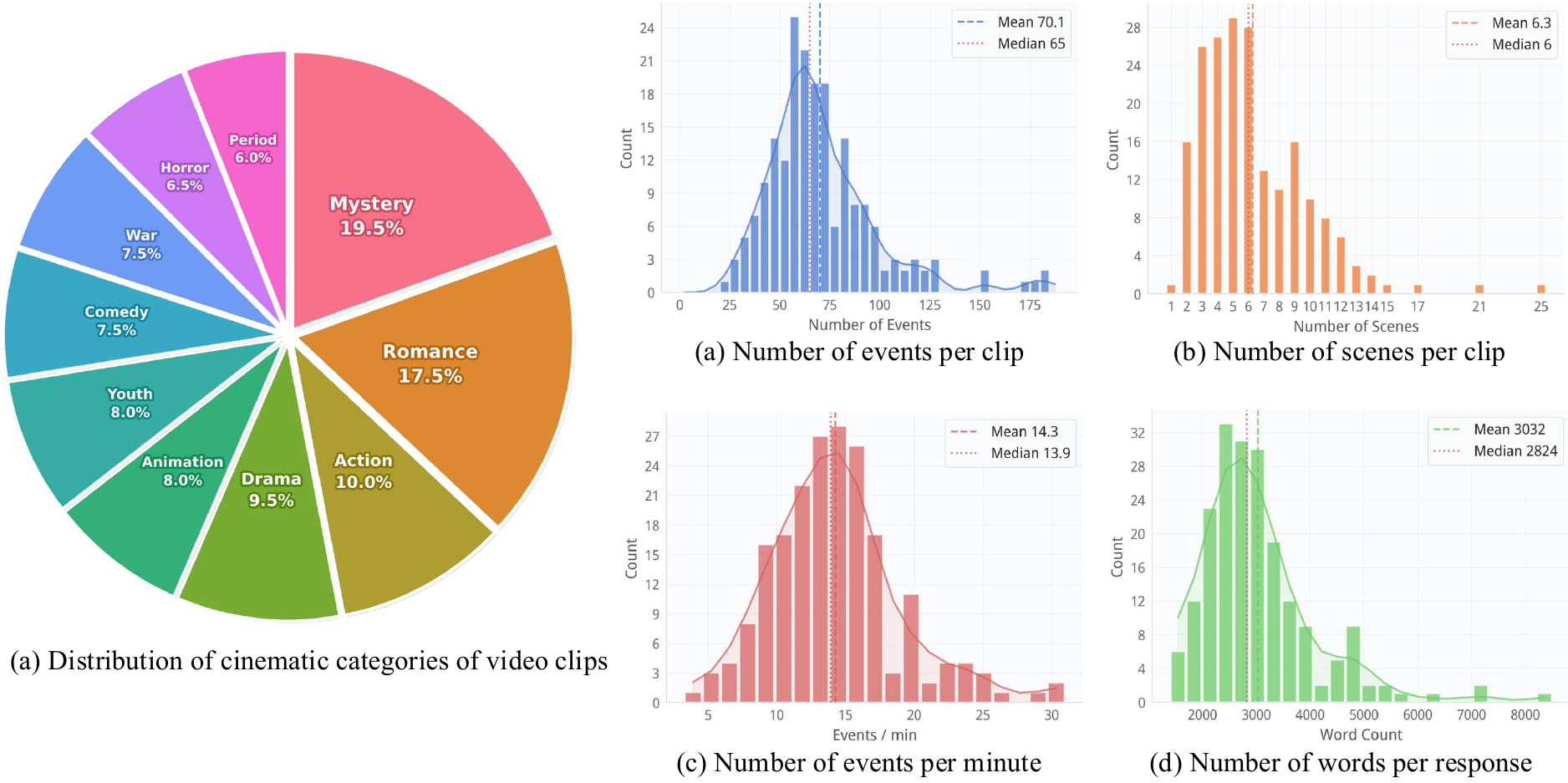}
    \caption{Statistics of 5-minute video clips in our benchmark.
    }
    \label{fig:bench_statistics}
    
\end{figure}

To comprehensively evaluate the capabilities of MLLMs in long-form video understanding and script generation, our benchmark is meticulously constructed to emphasize cinematic diversity, ultra-high annotation density, and linguistic richness. In this section, we present the statistical distributions of 5-minute video clips in our benchmark. For longer video clips (10-30 minutes) in our benchmark, the distribution of cinematic categories and number of events per minute are similar, and the number of words per response, the number of events per clip, and the number of scenes per clip increases proportionally.

\noindent \textbf{Cinematic Genre Diversity}
To ensure that models are evaluated on a robust and highly generalizable visual domain, our benchmark encompasses a wide array of cinematic categories. As illustrated in Figure~\ref{fig:bench_statistics} (a), the video clips are distributed across 10 distinct genres. While narrative-centric genres such as Mystery (19.5\%) and Romance (17.5\%) constitute the largest proportions, the dataset maintains a healthy, balanced long-tail distribution that includes Action (10.0\%), Drama (9.5\%), Animation (8.0\%), Youth (8.0\%), Comedy (7.5\%), War (7.5\%), Horror (6.5\%), and Period pieces (6.0\%). This structural diversity prevents models from overfitting to a specific lighting condition, shot composition, or narrative pacing, thereby rigorously testing their ability to generalize across drastically different cinematic styles.


\noindent \textbf{Granular Temporal Density and Structural Complexity}
A defining characteristic of our benchmark is its exceptionally high annotation density, which significantly departs from traditional video captioning datasets that typically provide sparse, high-level summaries. 
As shown in Figure~\ref{fig:bench_statistics}(b), the number of events per clip exhibits a dense distribution with a mean of 70.1 and a median of 65. This complexity is further compounded by the spatial-temporal transitions within the videos. Figure~\ref{fig:bench_statistics}(c) demonstrates that each clip contains an average of 6.3 distinct scenes, requiring models to accurately track storylines across multiple locations. 
Furthermore, the temporal granularity is extremely high. Figure~\ref{fig:bench_statistics}(d) reveals an average of 14.3 events annotated per minute (median 13.9), effectively translating to a distinct narrative action being grounded every 4 seconds. This extreme temporal density necessitates true continuous video reasoning. To succeed on this benchmark, models cannot rely on sparse keyframe sampling, they must exhibit fine-grained temporal perception to capture rapidly changing micro-actions, character interactions, and subtle scene boundary transitions without suffering from information loss.

\noindent \textbf{Linguistic Richness and Long-Context Challenge}
Beyond visual understanding, our benchmark pushes the boundaries of text generation length for video understanding tasks. As depicted in Fig.~\ref{fig:bench_statistics}(e), the total word count per generated script is massive, with a mean of 3,032 words and a median of 2,824 words, and a long tail extending up to nearly 8,000 words. 
This immense linguistic scale positions our dataset as an unprecedented challenge for long-context generation. It explicitly tests the memory retention and semantic coherence of MLLMs, challenging them to consistently maintain character identities, logical narrative flow, and professional script formatting over thousands of words, which are indispensable for real-world video script generation but remain largely unaddressed by existing short-form video benchmarks.

\end{document}